\documentclass[journal,10pt]{IEEEtran}
\usepackage{cite}
\usepackage{amsmath,amssymb,amsfonts}
\usepackage{algorithm, algorithmic}
\usepackage{graphicx}
\usepackage{subfigure}
\usepackage{textcomp}
\usepackage{multirow}
\usepackage{stfloats}
\usepackage{url}
\usepackage{color}
\usepackage{bm}
\usepackage{makecell}
\usepackage{flushend}
\usepackage{float}
\usepackage{threeparttable}
\usepackage{setspace}
\usepackage{upgreek}
\usepackage{bbm}
\usepackage{pifont}

\usepackage[colorlinks,
linkcolor=blue,
anchorcolor=blue,
citecolor=blue, 
urlcolor=black,
]{hyperref}

\newcommand{\nop}[1]{}
\allowdisplaybreaks[4]

\begin{document}
	
	\title{Joint Energy Management and Coordinated AIGC Workload Scheduling for Distributed Data Centers: A Diffusion-Aided Reward Shaping Approach}
	
	\author{~Yang~Fu,~Peng~Qin,~\IEEEmembership{Member,~IEEE}, Liming Chen, Zihao Zhang, Hao Yu, and Yifei Wang
		}
		
	\markboth{IEEE Transactions on Industrial Informatics}
	{}

	\maketitle

	\begin{abstract}	
     Artificial intelligence-generated content (AIGC) has emerged as a transformative paradigm for automating the creation of diverse and customized content, giving rise to rapidly growing computational workloads in cloud data centers. It is imperative for AIGC service providers (ASPs) to strategically schedule AIGC workloads to reduce data center energy costs while guaranteeing high-quality content generation. However, the distinctive characteristics of AIGC services pose critical challenges, including model heterogeneity across ASPs, implicit service quality evaluation, and complex inference process control. To tackle these challenges, we propose a joint energy management and coordinated AIGC workload scheduling framework, which introduces an explicit mathematical characterization of service quality to promote both job transfer among ASPs and fine-grained inference process configuration. Moreover, various energy resources within data centers are jointly considered to enhance power usage flexibility. Subsequently, a system utility maximization problem is formulated to balance AIGC service revenue with operational penalties and costs. Nevertheless, the strong coupling among job scheduling decisions induces severe reward sparsity, which limits the effectiveness of existing deep reinforcement learning (DRL) algorithms. To address this issue, we develop a diffusion model-aided reward shaping approach to synthesize complementary reward signals through a multi-step denoising process. This approach is seamlessly integrated with DRL to enable efficient learning of scheduling policies under sparse environmental feedback. Experiments based on real-world models and datasets demonstrate that our scheme effectively accommodates electricity price fluctuations and AIGC model heterogeneity, while achieving superior learning convergence and system utility compared with benchmark methods. 
	\end{abstract}
	
	
	\IEEEpeerreviewmaketitle
	
	\section{Introduction}
	
	\subsection{Background and Motivation}
	
	\IEEEPARstart{R}{ecent} breakthroughs in artificial intelligence-generated content (AIGC) have demonstrated unprecedented capabilities in automating the creation of diverse content spanning text, images, and videos \cite{1}. This transformative technology is reshaping modern society by enabling efficient and highly customized content generation tailored to various user demands \cite{2}. Meanwhile, the proliferation of AIGC services imposes substantial computational workloads on cloud data centers, resulting in rapidly escalating energy consumption and carbon emissions, e.g., ChatGPT processes about 2.5 billion prompts per day \cite{3} and generates over 260,000 kg of CO$_{\text{2}}$ monthly in 2025 \cite{4}. Therefore, how to strategically schedule AIGC workloads to reduce data center energy costs while guaranteeing high-quality generation services has become a pivotal problem for enhancing the sustainability of smart grids.
	
	Existing job scheduling schemes for distributed data centers exploits both spatial flexibility (by transferring jobs from data centers with high electricity prices to those with low prices \cite{5}) and temporal flexibility (by shifting workloads to low-price periods \cite{6}) to regulate power consumption patterns and decrease energy costs. However, the distinctive characteristics of AIGC services fundamentally distinguish them from traditional computing jobs, giving rise to several new challenges for AIGC workload scheduling:
	
	\textbf{$\bullet$ Heterogeneity of AIGC models:} AIGC models differ in architecture, parameter scale, and training dataset, leading to heterogeneous content generation capabilities, computational resource demands, and power consumption characteristics \cite{7}. Consequently, effective job scheduling requires selecting an appropriate AIGC service provider (ASP) by considering the attributes of models deployed at its data centers.
	
	\textbf{$\bullet$ Implicit evaluation of service quality:} Unlike traditional jobs that primarily emphasize completion delay, the quality of AIGC services is governed by the alignment between generated content and user preferences, while being coupled with data center energy consumption and operational costs. Nevertheless, the lack of a definitive mathematical formulation for service quality complicates the scheduling procedure \cite{8}. 
	
	\textbf{$\bullet$ Complexity in inference process control:} Existing schedulers typically determine only the execution location and timing of jobs, while AIGC workload entails fine-grained control over the model inference process. For instance, diffusion models rely on iterative denoising refinement to generate detailed content, and the number of denoising steps critically affects both content quality and service latency \cite{9,10}. 
	
	\subsection{Related Works}
	
	To date, research on AIGC workload scheduling among distributed data centers, aimed at balancing service quality and energy consumption, remains limited. Nonetheless, inspirations can be drawn from the following relevant directions. 
	
    \textit{1) Coordinated Job Scheduling:} This category focuses on optimizing job assignment and execution through coordinated strategies to reduce the total energy cost of distributed data centers. Work in \cite{11} proposed an electricity cost-aware job scheduling framework to achieve joint job sequencing and data center server selection, minimizing energy cost under deadline constraints. Reference \cite{12} developed a two-timescale deep reinforcement learning (DRL) approach, which schedules each job to an appropriate data center in the short-timescale and scales computational resource to adapt to long-term workload changes. \cite{13} designed a multi-objective job scheduler based on ensemble learning, enabling the intelligent search for the Pareto front of job completion delay and data center carbon emission. \cite{14} jointly optimized job scheduling and cooling regulation across distributed data centers to strike a balance among operational expenditure, waiting delay and power usage effectiveness. However, the aforementioned studies assume that full operational information of all data centers is globally available, enabling centralized job scheduling. This assumption is inconsistent with practical AIGC systems, where different ASPs retain exclusive ownership of their AIGC models, which cannot be accessed by others due to privacy concerns. Although distributed job scheduling has been explored in \cite{15} and \cite{16} using federated learning and the alternating direction method of multipliers, respectively, these approaches fail to capture the inherent heterogeneity of AIGC models across different ASPs. 

    \textit{2) Data Center Energy Management:} This category emphasizes optimizing power consumption by managing diverse flexible energy resources within data centers, such as computing servers, battery energy storage system (BESS), and renewable energy generation. Reference \cite{17} established a detailed energy management model incorporating multiple critical parts of electric loads, BESS, and power supply, with a hybrid quantum-Benders’ decomposition algorithm designed to optimize power usage. Work in \cite{18} proposed an adaptive power capping approach for data center energy management, which reduces both energy cost and deadline violation through learning the environment dynamics, mitigating the dependency on job-level information. \cite{19} developed a distributionally robust optimization framework, which leverages ambiguity set and column-and-constraint generation algorithm to minimize data center cost under uncertain renewable energy. The authors of \cite{20} adopted dynamic voltage and frequency scaling (DVFS) to reduce the server power consumption, while developing a reliability enhancement method to ensure the job deadline and energy constraints. \cite{21} proposed a joint day-ahead and intraday energy management solution to minimize the operational cost of data center micro grid. Nevertheless, existing schemes typically rely on abstract computational resource allocation or adjust only server frequencies, with limited consideration of GPU resource scaling. For AIGC workloads that are predominantly GPU-intensive, power consumption is influenced by multiple factors related to both GPU cores and memories, which further complicates data center energy management \cite{22}.
    
    \textit{3) AI Workload Scheduling:} This category investigates efficient scheduling of computation-intensive AI workloads in cloud data centers, encompassing model training, fine-tuning, and inference. Work in \cite{23} proposed a flow scheduler for AI training jobs to dynamically control the sending rates of tensors from each server, improving data center bandwidth utilization and accelerating the distributed learning process. Literature \cite{24} developed a carbon-efficient deep learning workload scheduler to carefully model the computing energy consumption and carbon footprints, with GPU allocation and frequency configuration optimized to minimize the job completion time. \cite{25} designed a multi-agent DRL algorithm to schedule model fine-tuning workloads across distributed data centers, striking a balance between job completion, cost reduction, and clear energy utilization. \cite{26} adopted DRL to optimize batch size, GPU core frequency, and memory frequency so as to minimize the inference energy consumption adhering to job delay constraints, with offline prediction method invoked to boost DRL preparedness. The authors of \cite{27} leveraged non-cooperative game to optimize the distribution of AI inference workloads across distributed data centers, decreasing the total operational costs and carbon emissions. Despite these advancements, these studies neither establish tailored performance metrics for AIGC workloads nor perform precise control of the inference process. AIGC services involve a complex interplay among content quality, user experience, completion delay, and energy consumption, necessitating dedicated modeling and system optimization \cite{28}. 
    
    \subsection{Contributions}
    
    In this paper, we investigate the joint energy management and coordinated AIGC workload scheduling for distributed data centers. We first propose a novel AIGC service metric that mathematically characterizes the impact of model attributes and key inference parameters on service quality. Subsequently, we develop a diffusion model-enhanced DRL framework to enable distributed AIGC workload scheduling without requiring access to the private model information of ASPs, while coordinating diverse computing and energy resources within data centers to enhance overall service utility. Our main contributions are summarized in the sequel. 
    
    \begin{enumerate}
		\item We propose an AIGC workload scheduling model for distributed data centers operated by different ASPs, which enables coordinated job transfer and fine-grained inference process configuration, while considering AIGC model heterogeneity, electricity price fluctuations, and renewable generation variability. In addition, multiple energy management components, including GPU DVFS, BESS charging/discharging, and cooling control, are incorporated to enhance power usage flexibility. Under this arrangement, a joint optimization problem is formulated to maximize system utility, which accounts for AIGC service revenue, deadline violation penalties, job transfer costs, and energy costs. 
		
		\item Owing to the strong coupling among job scheduling decisions as well as their collective impact on completion delay and generation result, the optimization problem exhibits severe reward sparsity that hampers the effective training of conventional DRL algorithms. To this end, we develop a diffusion model-aided reward shaping approach, which innovatively conditions the denoising process on state-action pairs to generate complementary reward signals, thereby enriching sparse environmental feedback and facilitating policy learning. Besides, we design an efficient heuristic with closed-form solutions for energy management optimization, which is embedded into the environment to form a DRL training loop.
		
		\item Extensive experiments based on real-world AIGC models, job traces, and electricity price data are conducted to evaluate the proposed approach and derive useful insights. Results demonstrate that our scheme strategically schedules AIGC workloads to respond to electricity price fluctuations, while adapting to the heterogeneous generation capabilities and computing overheads of different AIGC models. Moreover, the proposed reward shaping approach achieves up to a 1.5$\times$ improvement in cumulative reward over standard DRL algorithm, whilst the system utility is increased by more than 30\% compared with baseline schedulers that lack coordinated job transfer and inference process control.
	\end{enumerate}
    
    \textit{Remark 1:} In this work, diffusion models are considered as a representative class of AIGC models for image generation based on user prompts, which is consistent with their original design purpose \cite{2}. In contrast, for the proposed reward shaping approach, diffusion models are customized to synthesize complementary rewards that assist DRL in making job scheduling decisions. Although both employ a multi-step denoising principle, the diffusion model used for reward shaping is significantly more lightweight than that adopted for AIGC content generation.
    
	\section{System Model} \label{sec:model}

\begin{figure}[t]
    	\begin{center}
    		\centerline{\includegraphics[width=8.8cm]{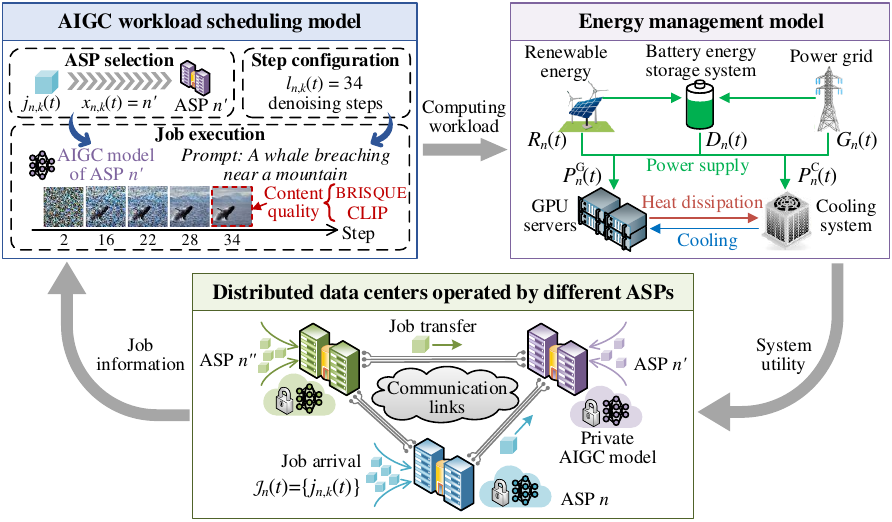}}
    	\end{center}
    	\vspace{-4mm}		
    	\caption{Joint energy management and AIGC workload scheduling framework for distributed data centers.} 
    	\label{fig:1}
    \end{figure}

	As illustrated in Fig. \ref{fig:1}, we consider a cloud-enabled AIGC service system comprising a set $\mathcal{N}=\left\{ 1,\ldots ,n,\ldots ,N \right\}$ of $N$ distributed data centers interconnected via high-speed cable communication links. Without loss of generality, each data center is supposed to be owned by a specific ASP, which employs its trained AIGC model to process incoming jobs while jointly managing diverse computing and energy resources within the data center to improve service utility. Moreover, the system’s operational horizon is discretized into $T$ time slots, denoted by $\mathcal{T}=\left\{ 1,\ldots ,t,\ldots ,T \right\}$, each with a duration of $\tau $. Detailed models and metrices are elucidated in the following subsections. 

	\subsection{AIGC Workload Scheduling Model}
	
	To facilitate modeling and analysis, we take diffusion model-based text-to-image generation as a representative AIGC service in this paper. Nevertheless, the proposed framework can be readily extended to other types of AIGC services (e.g., natural language processing) by substituting the associated resource consumption patterns and performance metrics. In the considered diffusion-based service, given a text prompt, the target image is generated from an initial random noise through a multi-step denoising process \cite{9}. The quality of the generated image, as well as the required computing workload, is significantly influenced by the adopted diffusion model and the number of denoising steps, which should be judiciously considered during job scheduling.
	
	In each time slot $t$, the jobs arrive at ASP $n$ is denoted by set ${{\mathcal{J}}_{n}}\left( t \right)=\left\{ {{j}_{n,k}}\left( t \right) \right\}$, where $k\in \left\{ 1,\ldots ,\left| {{\mathcal{J}}_{n}}\left( t \right) \right| \right\}$. Each job ${{j}_{n,k}}\left( t \right)$ is characterized by the text prompt ${{\mathbf{p}}_{n,k}}\left( t \right)$, tolerated delay ${{d}_{n,k}}\left( t \right)$, and transferability ${{\gamma }_{n,k}}\left( t \right)\in \left\{ 0,1 \right\}$. Since AIGC service provision corresponds to the model inference stage, which typically imposes real-time requirements, we have ${{d}_{n,k}}\left( t \right)<\tau$ and temporal job postponement is not permitted. Note that ${{d}_{n,k}}\left( t \right)$ is distinct for different jobs, if a job is not completed before its deadline, it is regarded as a failure. Besides, ${{\gamma }_{n,k}}\left( t \right)=1$ indicates that ${{j}_{n,k}}\left( t \right)$ can be transferred to the data center of another ASP for collaborative processing, otherwise it can only be processed by ASP $n$ for ${{\gamma }_{n,k}}\left( t \right)=0$. The value of ${{\gamma }_{n,k}}\left( t \right)$ may be determined by the computing environment reliance and user preference \cite{15}. 
	
	The job scheduling decision of ${{j}_{n,k}}\left( t \right)$ incorporates two variables. The first is ASP selection, given by
	\begin{align}
		{{x}_{n,k}}\left( t \right)=
		\left\{\begin{array}{cl}
			&\!\!\!\!\!\!\!\!\!{n}'\in \mathcal{N},\ {{\gamma }_{n,k}}\left( t \right)=1,\\
			&\!\!\!\!\!\!\!\!\!n,\ \ \,\qquad{{\gamma }_{n,k}}\left( t \right)=0,
		\end{array}\right. \label{eq:1}
	\end{align}
	which indicates that ${{j}_{n,k}}\left( t \right)$ is processed by the AIGC model of ASP ${{x}_{n,k}}\left( t \right)$. The second is the number of denoising steps ${{l}_{n,k}}\left( t \right)\in \mathcal{L}$, where $\mathcal{L}$ is the set of candidate steps with $L=\left| \mathcal{L} \right|$. Due to the private model ownership, ${{x}_{n,k}}\left( t \right)$ and ${{l}_{n,k}}\left( t \right)$ must be determined without accessing the internal models of other ASPs. With given scheduling result, the generated content for ${{j}_{n,k}}\left( t \right)$ is expressed as ${{\mathbf{c}}_{n,k}}\left( t \right)$.
	
	\subsection{Energy Management Model}
	
	In data centers, GPU servers and cooling systems constitute the primary sources of energy consumption. To reduce carbon emissions and enhance power supply flexibility, renewable energy and BESS are also incorporated.
	
	\subsubsection{GPU Servers}
	Unlike existing studies \cite{12,19,20} that simply adopt server frequency scaling models to characterize computing power consumption, we consider a GPU DVFS model tailored to AIGC workloads. Let $f_{n}^{\text{c}}\left( t \right)$, $V_{n}^{\text{c}}\left( t \right)$, and $f_{n}^{\text{m}}\left( t \right)$ denote the core frequency, core voltage, and memory frequency of ASP $n$’s GPU servers in time slot $t$, which are the major factors that impact the GPU power consumption $P_{n}^{\text{G}}\left( t \right)$ \cite{26}. Mathematically, we have \cite{29}
	\begin{align}
		P_{n}^{\text{G}}\left( t \right)=P_{n}^{0}+{{\lambda }_{n}}f_{n}^{\text{m}}\left( t \right)+{{\delta }_{n}}{{\left( V_{n}^{\text{c}}\left( t \right) \right)}^{2}}f_{n}^{\text{c}}\left( t \right), \label{eq:2}
	\end{align}
    where $P_{n}^{0}$ is the static power\footnote{Since the power consumption of the CPU in AIGC tasks is typically much lower than that of the GPU, it is treated as a constant and included in $P_{n}^{0}$.}, ${{\lambda }_{n}}$ and ${{\delta }_{n}}$ depend on the characteristics of GPU hardware and AIGC model. With DVFS parameters and scheduling results, the execution time of ${{j}_{n,k}}\left( t \right)$ can be calculated by\footnote{In data centers with heterogeneous GPU generations, servers can be partitioned into multiple virtual clusters, each consisting of homogeneous hardware and characterized by its own computing time model. Accordingly, job scheduling can be refined to select a specific cluster within an ASP, which only requires extending the action without altering the algorithm procedure.}
    \begin{align}
       \Delta _{n,k}^{\text{exe}}\!\left( t \right)\!=\!\!\!\!\sum\limits_{{n}'\in \mathcal{N}}\!\!{{{\mathbb{I}}_{\left\{ {{x}_{n,k}}\left( t \right)={n}' \right\}}}{{l}_{n,k}}\!\left( t \right)\!\!\left[ \Delta _{{{n}'}}^{0}\!\!+\!{{\varepsilon }_{{{n}'}}}\!\left(\! \frac{{{\kappa }_{{{n}'}}}}{f_{{{n}'}}^{\text{c}}\!\left( t \right)}\!\!+\!\!\frac{1\!\!-\!{{\kappa }_{{{n}'}}}}{f_{{{n}'}}^{\text{m}}\!\left( t \right)} \!\right) \!\right]}\!, \label{eq:3}
    \end{align}
    where ${{\mathbb{I}}_{\left\{ \cdot  \right\}}}$ is an indicator function with ${{\mathbb{I}}_{\left\{ h \right\}}}=1$ when $h$ is true, otherwise ${{\mathbb{I}}_{\left\{ h \right\}}}=0$. $\Delta _{n}^{0}$ specifies the basic time for executing one denoising step, ${{\varepsilon }_{n}}$ and ${{\kappa }_{n}}$ imply the time sensitivity to DVFS. The correlation between core frequency and voltage is characterized by a sublinear function \cite{29}: $f_{n}^{\text{c}}\left( t \right)\le \sqrt{{\left[ V_{n}^{\text{c}}\left( t \right)-{{\chi }_{n}} \right]}/{2}}+{{\chi }_{n}}\triangleq h\left[ V_{n}^{\text{c}}\left( t \right) \right]$ with ${{\chi }_{n}}$ being a constant, implying that the maximum allowed $f_{n}^{\text{c}}\left( t \right)$ is decided by $V_{n}^{\text{c}}\left( t \right)$. Note that the above constant coefficients can be estimated by running the AIGC model on the GPU servers of each ASP $n$ in offline. 
    
    \subsubsection{Cooling System}
    To ensure the normal operation of servers, cooling system with computer room air conditioning (CRAC) units is utilized to maintain suitable temperature of data center. According to the thermal energy balance equation, we have \cite{30}
    \begin{align}
    	& P_{n}^{\text{G}}\!\left( t \right)\!-\!{{\vartheta }^{\text{COP}}}\! P_{n}^{\text{C}}\!\left( t \right)\!+\!\frac{\zeta _{n}^{\text{out}}\!\left( t \right)\!-\!\zeta _{n}^{\text{in}}\!\left( t \right)}{{{\Omega }_{n}}}\!=\!\rho c{{V}_{n}}\frac{\zeta _{n}^{\text{in}}\!\left( t\! +\!1 \right)\!-\!\zeta _{n}^{\text{in}}\!\left( t \right)}{\tau },\nonumber\\
    	&\qquad\qquad\qquad\qquad\qquad\qquad\qquad\qquad\qquad\qquad\forall n,t, \label{eq:4}
    \end{align}
    where the heat dissipation power of servers approximately equals to $P_{n}^{\text{G}}\left( t \right)$. ${{\vartheta }^{\text{COP}}}P_{n}^{\text{C}}\left( t \right)$ represents the cooling power with $P_{n}^{\text{C}}\left( t \right)$ being the electric power consumption of the cooling system and ${{\vartheta }^{\text{COP}}}$ the coefficient of performance (COP)\footnote{Similar to \cite{5,6,21}, the COP of the CRAC system is modeled as a constant to represent its average operating efficiency. This commonly adopted abstraction enables us to focus on AIGC workload scheduling, while a dynamic COP can be incorporated without altering our optimization approach.}.. $\zeta _{n}^{\text{out}}\left( t \right)$ and $\zeta _{n}^{\text{in}}\left( t \right)$ are the outside and inside temperatures in slot $t$, respectively, ${{\Omega }_{n}}$ denotes the thermal resistance. $\rho$ and $c$ are the density and specific heat capacity of air, respectively, ${{V}_{n}}$ indicates the data center volume. The inside temperature and temperature change should satisfy the following constraints:  
    \begin{align}
    	&\qquad \zeta _{n}^{\text{in},\min }\le \zeta _{n}^{\text{in}}\left( t \right)\le \zeta _{n}^{\text{in},\max },\ \forall n,t,\label{eq:5}\\
    	&\left| \zeta _{n}^{\text{in}}\left( t+1 \right)-\zeta _{n}^{\text{in}}\left( t \right) \right|\le \Delta \zeta _{n}^{\text{in},\max },\ \forall n,t, \label{eq:6}
    \end{align}
	where $\zeta _{n}^{\text{in},\min },\zeta _{n}^{\text{in},\max }$ are the bounds of data center temperature, $\Delta \zeta _{n}^{\text{in},\max }$ is the maximum temperature change. 
	
	\subsubsection{Renewable Energy and BESS}
	Each data center deploys a BESS for storing the energy from renewable sources and power grid. Denote the remaining energy of BESS at ASP $n$ in slot $t$ as ${{E}_{n}}\left( t \right)$, ${{D}_{n}}\left( t \right)$ indicates the discharging/charging power (${{D}_{n}}\left( t \right)>0$ when discharging and ${{D}_{n}}\left( t \right)<0$ when charging), then we have
	\begin{align}
		&\quad\qquad\ {{E}_{n}}\left( t+1 \right)={{E}_{n}}\left( t \right)-{{D}_{n}}\left( t \right)\tau ,\ \forall n,t,\label{eq:7}\\
		&\quad\qquad\qquad\quad0\le {{E}_{n}}\left( t \right)\le E_{n}^{\max },\ \forall n,t, \label{eq:8}\\
		&-\min \left\{ D_{n}^{\text{min}},{\left[ E_{n}^{\max }-{{E}_{n}}\left( t \right) \right]}/{\tau } \right\}\le {{D}_{n}}\left( t \right)\nonumber\\
		&\qquad\qquad\qquad\qquad\le \min \left\{ D_{n}^{\text{max}},{{{E}_{n}}\left( t \right)}/{\tau } \right\},\ \forall n,t, \label{eq:9}
	\end{align}
    where $E_{n}^{\max }$ signifies the capacity of BESS, $D_{n}^{\text{max}}>0$ and $D_{n}^{\text{min}}>0$ are the maximum discharging and charging power, respectively. Accordingly, the electric power balance is represented as 
    \begin{align}
    	P_{n}^{\text{G}}\left( t \right)+P_{n}^{\text{C}}\left( t \right)={{D}_{n}}\left( t \right)+{{R}_{n}}\left( t \right)+{{G}_{n}}\left( t \right),\ \forall n,t, \label{eq:10}
    \end{align}
    where ${{R}_{n}}\left( t \right)$ denotes the renewable power generated at ASP $n$. ${{G}_{n}}\left( t \right)$ indicates the power exchange between the data center and the grid, ${{G}_{n}}\left( t \right)>0$ means that the data center absorbs energy from the grid, ${{G}_{n}}\left( t \right)<0$ means that the electricity is sold back to the grid. 
    
	\subsection{System Utility}
	
	The system utility accounts for both revenues and negative costs during the AIGC service procedure. Specifically, we introduce four utility components in the sequel. 
	\begin{figure}[t]
		\begin{center}
			\centerline{\includegraphics[width=6cm]{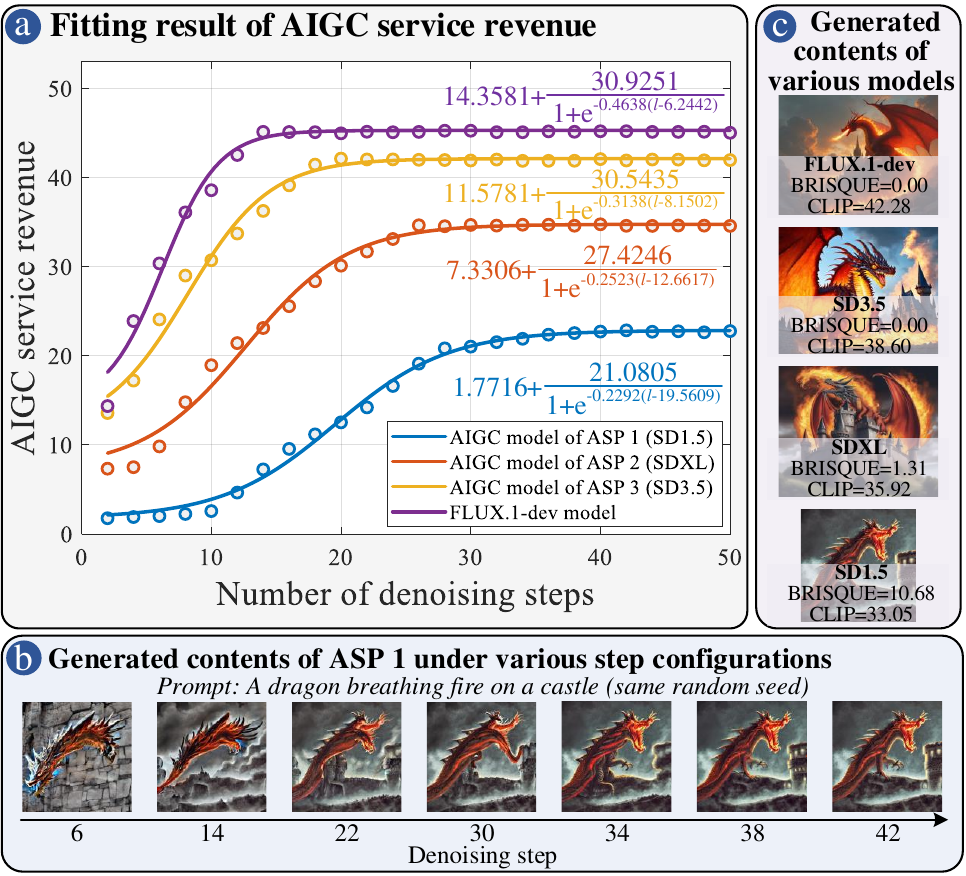}}
		\end{center}
		\vspace{-4mm}		
		\caption{Illustrations of AIGC service revenue.} 
		\label{fig:2}
	\end{figure}  

	\subsubsection{AIGC Service Revenue}
	In the context of AIGC services, an ASP’s revenue is strongly influenced by the quality of the content generated by its diffusion model. To capture this effect, we design a tailored evaluation metric that integrates both objective image quality and subjective prompt alignment. First, blind/referenceless image spatial quality evaluator (BRISQUE) assesses a score, which is negatively correlated to the objective image quality, based on various perception factors such as contrast, sharpness, and texture, without requiring a reference image\footnote{\url{http://live.ece.utexas.edu/research/quality/BRISQUE_release.zip}}. Second, contrastive language-image pre-training (CLIP) leverages vision language models to measure semantic alignment between generated image and text prompt, reflecting subjective prompt adherence\footnote{https://github.com/openai/CLIP}. As a result, the AIGC service revenue for completing job ${{j}_{n,k}}\left( t \right)$ is expressed as
	\begin{align}
		\!\!\hat{U}_{n,k}^{\text{R}}\!\left( t \right)\!=\!-\text{BRISQUE}\!\left( {{\mathbf{c}}_{n,k}}\!\left( t \right) \right)\!+\!\text{CLIP}\!\left( {{\mathbf{c}}_{n,k}}\!\left( t \right)\!,{{\mathbf{p}}_{n,k}}\!\left( t \right) \right)\!.\!\! \label{eq:11}
	\end{align}
    To gain further insights, we conduct extensive experiments using different diffusion models, as illustrated in Fig. \ref{fig:2}, to evaluate $\hat{U}_{n,k}^{\text{R}}\left( t \right)$ under varying numbers of denoising steps. Detailed setup will be provided in Section V. Numerical results reveal that $\hat{U}_{n,k}^{\text{R}}\left( t \right)$ increases and eventually stabilizes as the number of denoising steps grows, while different models exhibit distinct growth rates and stabilized values. Motivated by these observations, we fit $\hat{U}_{n,k}^{\text{R}}\left( t \right)$ employing a modified sigmoid function: 
    \begin{align}
    	U_{n,k}^{\text{R}}\left( t \right)\!\!=\!\!\!\sum\limits_{{n}'\in \mathcal{N}}\!\!{{{\mathbb{I}}_{\left\{ {{x}_{n,k}}\left( t \right)={n}' \right\}}}\!\!\left\{\! {{\xi }_{{{n}'}}}\!+\!\frac{{{{\tilde{\xi }}}_{{{n}'}}}}{1\!+\!{{\text{e}}^{-{{{\bar{\xi }}}_{{{n}'}}}\left[ {{l}_{n,k}}\left( t \right)\!-\!{{{\hat{\xi }}}_{{{n}'}}} \right]}}}\! \right\}}, \label{eq:12}
    \end{align}
    where ${{\xi }_{n}},{{\bar{\xi }}_{n}},{{\tilde{\xi }}_{n}},{{\hat{\xi }}_{n}}$ characterize both the maximum achievable content quality and the sensitivity to the number of denoising steps. Different AIGC models possess distinct sets of fitting parameters, reflecting model heterogeneity across ASPs\footnote{While this work focuses on text-to-image generation as a representative AIGC task, our methodology for modeling the service revenue function can be readily extended to other tasks, e.g., video generation and high-resolution synthesis. In these cases, similar experiments can be conducted to quantify the impact of controllable inference parameters on generation quality. The fitted functions can then be substituted into (\ref{eq:12}) for job scheduling optimization. }.
    
    \subsubsection{Deadline Violation Penalty}
    The job completion delay stems from job transfer, waiting, execution, and content feedback. Define $s\left[ {{\mathbf{p}}_{n,k}}\left( t \right) \right]$ as the data size of job ${{j}_{n,k}}\left( t \right)$’s prompt, the transfer delay is calculated by
    \begin{align}
    	\Delta _{n,k}^{\text{tra}}\left( t \right)=\sum\limits_{{n}'\in \mathcal{N}\backslash \left\{ n \right\}}{{{\mathbb{I}}_{\left\{ {{x}_{n,k}}\left( t \right)={n}' \right\}}}\frac{s\left[ {{\mathbf{p}}_{n,k}}\left( t \right) \right]}{{{r}_{n,{n}'}}\left( t \right)}}, \label{eq:13}
    \end{align}
    where ${{r}_{n,{n}'}}\left( t \right)$ denotes the transmission rate between ASP $n$ and ${n}'$ in slot $t$. The waiting delay accounts for the average sojourn time among all jobs scheduled to the selected ASP ${{x}_{n,k}}\left( t \right)$ until ${{j}_{n,k}}\left( t \right)$ is processed, given by
    \begin{align}
    	\Delta _{n,k}^{\text{wai}}\!\left( t \right)\!=\!\frac{1}{2}\!\underbrace{\sum\limits_{{{j}_{{n}',{k}'}}\left( t \right)\in \mathcal{J}\left( t \right)\backslash \left\{ {{j}_{n,k}}\left( t \right) \right\}}\!\!{{{\mathbb{I}}_{\left\{ {{x}_{n,k}}\left( t \right)={{x}_{{n}',{k}'}}\left( t \right) \right\}}}\!\frac{\Delta _{{n}',{k}'}^{\text{exe}}\!\left( t \right)}{{{\Theta }_{{{x}_{n,k}}\left( t \right)}}}}}_{\left( \text{I} \right)}, \label{eq:14}
    \end{align}
    where $\mathcal{J}\left( t \right)={{\cup }_{n\in \mathcal{N}}}{{\mathcal{J}}_{n}}\left( t \right)$ is the set of all jobs in slot $t$, $\Delta _{{n}',{k}'}^{\text{exe}}\left( t \right)$ is the execution delay of ${{j}_{{n}',{k}'}}\left( t \right)$, as in (\ref{eq:3}), ${{\Theta }_{n}}$ denotes the number of GPUs at ASP $n$. The multiplication by $\frac{1}{2}$ indicates taking the expectation between zero waiting delay (${{j}_{n,k}}\left( t \right)$ arrives ahead of all jobs on the same GPU) and waiting delay of $\left( \text{I} \right)$ (${{j}_{n,k}}\left( t \right)$ arrives behind all jobs on the same GPU). Let $s\left[ {{\mathbf{c}}_{n,k}}\left( t \right) \right]$ be the data size of the generated content, the feedback delay of ${{j}_{n,k}}\left( t \right)$ is 
    \begin{align}
    	\Delta _{n,k}^{\text{fee}}\left( t \right)=\sum\limits_{{n}'\in \mathcal{N}\backslash \left\{ n \right\}}{{{\mathbb{I}}_{\left\{ {{x}_{n,k}}\left( t \right)={n}' \right\}}}\frac{s\left[ {{\mathbf{c}}_{n,k}}\left( t \right) \right]}{{{r}_{{n}',n}}\left( t \right)}}. \label{eq:15}
    \end{align}
    Therefore, the total completion delay of ${{j}_{n,k}}\left( t \right)$ is
    \begin{align}
    	{{\Delta }_{n,k}}\left( t \right)=\Delta _{n,k}^{\text{tra}}\left( t \right)+\Delta _{n,k}^{\text{wai}}\left( t \right)+\Delta _{n,k}^{\text{exe}}\left( t \right)+\Delta _{n,k}^{\text{fee}}\left( t \right). \label{eq:16}
    \end{align}
    Given job tolerated delay ${{d}_{n,k}}\left( t \right)$, the deadline violation penalty can be calculated by
    \begin{align}
    	C_{n,k}^{\text{D}}\left( t \right)={{\mathbb{I}}_{\left\{ {{\Delta }_{n,k}}\left( t \right)>{{d}_{n,k}}\left( t \right) \right\}}}\upsilon, \label{eq:17}
    \end{align}
    where $\upsilon$ denotes the penalty factor. 
    
    \subsubsection{Job Transfer Cost}
    The transfer of job ${{j}_{n,k}}\left( t \right)$ occupies of ASP’s switches and transmission bandwidth, leading to the following cost: 
    \begin{align}
    	C_{n,k}^{\text{T}}\left( t \right)={{\mathbb{I}}_{\left\{ {{x}_{n,k}}\left( t \right)\ne n \right\}}}\left\{ s\left[ {{\mathbf{p}}_{n,k}}\left( t \right) \right]+s\left[ {{\mathbf{c}}_{n,k}}\left( t \right) \right] \right\}\psi, \label{eq:18}
    \end{align}
    where $\psi$ represents the cost for transferring per bit data. 
    
    \subsubsection{Energy Cost}
    Recall that the power exchange between ASP $n$ and the grid is ${{G}_{n}}\left( t \right)$, the energy cost is given by
    \begin{align}
    	C_{n}^{\text{E}}\left( t \right)={{\varsigma }_{n}}\left( t \right){{G}_{n}}\left( t \right)\tau, \label{eq:19}
    \end{align}
    where ${{\varsigma }_{n}}\left( t \right)$ denotes the electricity price at ASP $n$ in slot $t$. 
    
    To conclude, the system utility in slot $t$ is calculated by
    \begin{align}
    	U\!\left( t \right)\!=\!\!\!\!\!\!&\sum\limits_{{{j}_{n,k}}\left( t \right)\in \mathcal{J}\left( t \right)}\!\!\!\!\!\!\!{\left[ U_{n,k}^{\text{R}}\!\left( t \right)\!{{\mathbb{I}}_{\left\{ {{\Delta }_{n,k}}\left( t \right)\le {{d}_{n,k}}\left( t \right) \right\}}}\!-\! C_{n,k}^{\text{D}}\!\left( t \right)\!-\! C_{n,k}^{\text{T}}\!\left( t \right) \right]}\nonumber\\
    	&\ -\sum\limits_{n\in \mathcal{N}}{C_{n}^{\text{E}}\left( t \right)}. \label{eq:20}
    \end{align}
    
    \section{Problem Formulation and Decomposition}
    
    \subsection{Problem Formulation}
    
    We formulate a joint energy management and AIGC workload scheduling problem, aiming to maximize the system utility over all time slots while satisfying a series of operational constraints. The optimization variables include ASP selection $\mathbf{x}\left( t \right)=\left\{ {{x}_{n,k}}\left( t \right):\forall n,k \right\}$, denoising step configuration $\mathbf{l}\left( t \right)=\left\{ {{l}_{n,k}}\left( t \right):\forall n,k \right\}$, DVFS of GPU servers $\mathbf{f}\left( t \right)=\left\{ f_{n}^{\text{c}}\left( t \right),V_{n}^{\text{c}}\left( t \right),f_{n}^{\text{m}}\left( t \right)\!:\!\forall n \right\}$, and power usage behavior $\mathbf{b}\left( t \right)=\left\{ P_{n}^{\text{C}}\left( t \right),{{D}_{n}}\left( t \right)\!:\!\forall n \right\}$. The problem is given by
    \begin{subequations}
    	\begin{equation}
    		\begin{aligned}
    			\textbf{P1}: \underset{\left\{ \mathbf{x}\left( t \right),\mathbf{l}\left( t \right),\mathbf{f}\left( t \right),\mathbf{b}\left( t \right):\forall t \right\}}{\mathop{\max }}\,\sum\limits_{t\in \mathcal{T}}{U\left( t \right)}, \label{eq:21a}
    		\end{aligned}
    	\end{equation}
    	\vspace{-5mm}
    	\begin{align}
    		\mbox{s.t.}\ 
    		&{{x}_{n,k}}\left( t \right)=
    		\left\{\begin{array}{cl}
    			&\!\!\!\!\!\!\!\!\!{n}'\in \mathcal{N},\ {{\gamma }_{n,k}}\left( t \right)=1,\\
    			&\!\!\!\!\!\!\!\!\!n,\ \ \,\qquad{{\gamma }_{n,k}}\left( t \right)=0,
    		\end{array}\right. \forall n,k,t, \label{eq:21b}\\
    		&{{l}_{n,k}}\left( t \right)\in \mathcal{L},\ \forall n,k,t, \label{eq:21c}\\
    		&\sum\limits_{{{j}_{n,k}}\left( t \right)\in {{\mathcal{J}}_{n}}\left( t \right)}{{{\mathbb{I}}_{\left\{ {{x}_{n,k}}\left( t \right)\ne n \right\}}}}\le J_{n}^{\max },\ \forall n,t, \label{eq:21d}\\  
    		&V_{n}^{\text{c},\min }\le V_{n}^{\text{c}}\left( t \right)\le V_{n}^{\text{c},\max },\forall n,t, \label{eq:21e}\\
    		&f_{n}^{\text{c},\min }\le f_{n}^{\text{c}}\left( t \right)\le h\left[ V_{n}^{\text{c}}\left( t \right) \right],\ \forall n,t, \label{eq:21f}\\
    		&f_{n}^{\text{m},\min }\le f_{n}^{\text{m}}\left( t \right)\le f_{n}^{\text{m},\max },\ \forall n,t, \label{eq:21g}\\
    		&0\le P_{n}^{\text{C}}\left( t \right)\le P_{n}^{\text{C},\max },\ \forall n,t, \label{eq:21h}\\
    		&\text{(\ref{eq:4})-(\ref{eq:10})}, \nonumber
    	\end{align} 
    \end{subequations}
    where (\ref{eq:21b}) and (\ref{eq:21c}) specify the definition of job scheduling variables. (\ref{eq:21d}) means that the number of transferred jobs at each ASP $n$ is no larger than the upper bound $J_{n}^{\max }$ due to communication capacity limits. (\ref{eq:21e})-(\ref{eq:21g}) restrict the DVFS of GPU servers\footnote{When practical GPU DVFS operates with discrete power states, the optimized values in $\mathbf{f}\left( t \right)$ can be quantized to their nearest discrete levels. Our work assumes sufficient DVFS granularity with negligible quantization loss, while coarse DVFS levels may introduce performance degradation.}, where $V_{n}^{\text{c},\min },V_{n}^{\text{c},\max }$ denote the maximum and minimum core voltages, $f_{n}^{\text{c},\min }$ is the minimum core frequency, $f_{n}^{\text{m},\min },f_{n}^{\text{m},\max }$ represent the maximum and minimum memory frequencies. In (\ref{eq:21h}), $P_{n}^{\text{C},\max }$ is the maximum power consumption of  the cooling system, respectively. (\ref{eq:4})-(\ref{eq:10}) incorporates constraints related to data center temperature, BESS operation, and electric power balance.  
    
    However, \textbf{P1} is an intractable mixed-integer nonlinear programming (MINLP) problem due to the presence of both discrete variables $\mathbf{x}\left( t \right),\mathbf{l}\left( t \right)$ and continuous variables $\mathbf{f}\left( t \right),\mathbf{b}\left( t \right)$. Besides, the update rules for the energy storage state and data center temperature in (\ref{eq:4}) and (\ref{eq:7}) indicate that these variables are temporally coupled, and such coupling is further exacerbated by unpredictable future job arrivals and renewable energy generation. In addition, owing to privacy requirements, each ASP must determine its optimization decisions in a distributed manner without access to the global information of other ASPs. 
    
    \subsection{Problem Decomposition}
    
    To address these challenges, we observe that the job scheduling variables dominate GPU DVFS and power usage behavior. Once the values of $\mathbf{x}\left( t \right),\mathbf{l}\left( t \right)$ are fixed, the achievable content generation quality of each job and the workload of each ASP become determined. Consequently, the optimization of $\mathbf{f}\left( t \right),\mathbf{b}\left( t \right)$ can be decoupled across ASPs, enabling each ASP to independently maximize its own utility while collectively contributing to the global optimum. 
    
    With this in mind, we decompose the original problem \textbf{P1} into two nested subproblems. The \textit{first outer-layer subproblem} optimizes $\mathbf{x}\left( t \right),\mathbf{l}\left( t \right)$ over time slots, while treating the implicit influence of $\mathbf{f}\left( t \right),\mathbf{b}\left( t \right)$ as part of the environmental feedback. This subproblem is formulated as 
    \begin{equation}
    	\begin{aligned}
    		\textbf{SP1}: \underset{\left\{ \mathbf{x}\left( t \right),\mathbf{l}\left( t \right):\forall t \right\}}{\mathop{\max }}\,\sum\limits_{t\in \mathcal{T}}{U\left( t \right)}, \label{eq:22}
    	\end{aligned}
    \end{equation}
    \vspace{-5mm}
    \begin{align}
    	\mbox{s.t.}\ \text{(\ref{eq:21b})-(\ref{eq:21d})}, \nonumber
    \end{align} 
    
    Given $\mathbf{x}\left( t \right),\mathbf{l}\left( t \right)$ in each time slot $t$, the \textit{second inner-layer subproblem} optimizes ${{\mathbf{f}}_{n}}\left( t \right)=\left\{ f_{n}^{\text{c}}\left( t \right),V_{n}^{\text{c}}\left( t \right),f_{n}^{\text{m}}\left( t \right) \right\}$ and ${{\mathbf{b}}_{n}}\left( t \right)=\left\{ P_{n}^{\text{C}}\left( t \right),{{D}_{n}}\left( t \right) \right\}$ for each ASP $n$. To cope with the temporal coupling in constraints (\ref{eq:4}) and (\ref{eq:7}), since $\zeta _{n}^{\text{in}}\left( t \right),{{E}_{n}}\left( t \right)$ at the beginning of $t$ are fixed, we convert (\ref{eq:4})-(\ref{eq:6}) to restrict $\zeta _{n}^{\text{in}}\left( t+1 \right)$ as follows
    \begin{align}
    	&\max \left\{ \tilde{\zeta }_{n}^{\min }\left( t \right),\Delta \tilde{\zeta }_{n}^{\min }\left( t \right) \right\}\le P_{n}^{\text{G}}\left( t \right)-{{\vartheta }^{\text{COP}}}P_{n}^{\text{C}}\left( t \right)\nonumber\\
    	&\qquad\qquad\qquad\qquad\le \min \left\{ \tilde{\zeta }_{n}^{\max }\left( t \right),\Delta \tilde{\zeta }_{n}^{\max }\left( t \right) \right\}, \label{eq:23}
    \end{align}
    where $\tilde{\zeta }_{n}^{\min }\left( t \right)=\frac{\rho c{{V}_{n}}}{\tau }\zeta _{n}^{\text{in},\min }-\left( \frac{\rho c{{V}_{n}}}{\tau }-\frac{1}{{{\Omega }_{n}}} \right)\zeta _{n}^{\text{in}}\left( t \right)-\frac{\zeta _{n}^{\text{out}}\left( t \right)}{{{\Omega }_{n}}}$, $\tilde{\zeta }_{n}^{\max }\left( t \right)=\frac{\rho c{{V}_{n}}}{\tau }\zeta _{n}^{\text{in},\max }-\left( \frac{\rho c{{V}_{n}}}{\tau }-\frac{1}{{{\Omega }_{n}}} \right)\zeta _{n}^{\text{in}}\left( t \right)-\frac{\zeta _{n}^{\text{out}}\left( t \right)}{{{\Omega }_{n}}}$, $\Delta \tilde{\zeta }_{n}^{\min }\left( t \right)=-\frac{\rho c{{V}_{n}}}{\tau }\Delta \zeta _{n}^{\text{in},\max }+\frac{\zeta _{n}^{\text{in}}\left( t \right)-\zeta _{n}^{\text{out}}\left( t \right)}{{{\Omega }_{n}}}$, and $\Delta \tilde{\zeta }_{n}^{\max }\left( t \right)=\frac{\rho c{{V}_{n}}}{\tau }\Delta \zeta _{n}^{\text{in},\max }+\frac{\zeta _{n}^{\text{in}}\left( t \right)-\zeta _{n}^{\text{out}}\left( t \right)}{{{\Omega }_{n}}}$. Moreover, in light of Lyapunov optimization, we introduce a BESS energy queue ${{\tilde{E}}_{n}}\left( t \right)=E_{n}^{\max }-{{E}_{n}}\left( t \right)$ to represent the amount of discharged energy. To strike a balance between reserving energy for future uncertainties and reducing the immediate energy cost, an additional term $-\Upsilon {{\tilde{E}}_{n}}\left( t \right){{D}_{n}}\left( t \right)\tau $, weighted by parameter $\Upsilon $, is incorporated into the objective function. The detailed derivation is analogous to that in \cite{32}, and is thus omitted here. This subproblem is formulated as
    \begin{align}
    	\textbf{SP2}_n(t): \underset{{{\mathbf{f}}_{n}}\left( t \right),{{\mathbf{b}}_{n}}\left( t \right)}{\mathop{\max }}&\sum\limits_{k\in \mathcal{K}_{n}^{*}\left( t \right)}{\left[ U_{k}^{\text{R}}\left( t \right){{\mathbb{I}}_{\left\{ {{\Delta }_{k}}\left( t \right)\le {{d}_{k}}\left( t \right) \right\}}}-C_{k}^{\text{D}}\left( t \right) \right]}\nonumber\\
    		&-C_{n}^{\text{E}}\left( t \right)-\Upsilon {{\tilde{E}}_{n}}\left( t \right){{D}_{n}}\left( t \right)\tau, \label{eq:24}
    \end{align}
    \vspace{-5mm}
    \begin{align}
    	\mbox{s.t.}\ \text{(\ref{eq:23}), (\ref{eq:21e})-(\ref{eq:21h}), and (\ref{eq:9})-(\ref{eq:10}) evaluated at }(n,t), \nonumber
    \end{align} 
    According to the optimized $\mathbf{x}\left( t \right)$, $\mathcal{K}_{n}^{*}\left( t \right)$ denotes the set of jobs that are scheduled to ASP $n$ in slot $t$, and the job index is simplified to $k$ for brevity. 
    
    \section{Joint Energy Management and AIGC Workload Scheduling Solution} 
    
    In this section, we present the solution methodology for the two subproblems described above. For \textbf{SP1}, we reformulate it as a Markov decision process (MDP) and propose a diffusion model-aided reward shaping approach, which is seamlessly integrated with DRL to enable distributed job scheduling under low-quality environmental feedback. Subsequently, \textbf{SP2}$_{n}\left( t \right)$ is efficiently solved using a heuristic method along with closed-form solutions.
    
    \subsection{MDP Reformulation for \textbf{SP1}}
    
    Since job scheduling must be carried out without global visibility, DRL offers an interactive learning framework that enables each ASP to infer the behavioral patterns of others, thereby facilitating distributed and coordinated decision-making. To align \textbf{SP1} with the DRL paradigm, we reformulate it as an MDP that makes sequential scheduling decisions for each job. Main elements are specified in the sequel. 
    
    \textit{1) State:} When a job ${{j}_{n,k}}\left( t \right)$ arrives at ASP $n$, the ASP acts as an agent to observe a state ${{\mathbf{o}}_{n,k}}\left( t \right)$ related to job information, communication condition, renewable power, electricity price, and the scheduling status of previous jobs, expressed as
    \begin{align}
    	&{{\mathbf{o}}_{n,k}}\left( t \right)=[ s\left[ {{\mathbf{p}}_{n,k}}\left( t \right) \right],{{d}_{n,k}}\left( t \right),{{\gamma }_{n,k}}\left( t \right),\nonumber\\
    	&\quad\qquad\left\{ {{r}_{n,{n}'}}\left( t \right),{{R}_{{{n}'}}}\left( t \right),{{\varsigma }_{{{n}'}}}\left( t \right),{{\Psi }_{{{n}'}}}\left( t \right):{n}'\in \mathcal{N} \right\} ], \label{eq:25}
    \end{align}
    where ${{\Psi }_{{{n}'}}}\left( t \right)$ indicates the AIGC workload that have been scheduled to ASP ${n}'$, which equals zero at the beginning of slot $t$ and gradually increases with the scheduling procedure. 
    
   \textit{2) Action:} Based on the observed state, action is made to jointly optimize ASP selection ${{x}_{n,k}}\left( t \right)$ and denoising step configuration ${{l}_{n,k}}\left( t \right)$. To make it more suitable for DRL output, we design the action as a $N+L$-dimensional vector
    \begin{align}
    	{{\mathbf{a}}_{n,k}}\!\left( t \right)\!=\!\left[ \tilde{x}_{n,k}^{\left( 1 \right)}\!\left( t \right),\ldots ,\tilde{x}_{n,k}^{\left( N \right)}\!\left( t \right),\tilde{l}_{n,k}^{\left( 1 \right)}\!\left( t \right),\ldots ,\tilde{l}_{n,k}^{\left( L \right)}\!\left( t \right) \right], \label{eq:26}
    \end{align}
    in which the first $N$ elements mean the probabilities for selecting $N$ ASPs, and the $N+1$ to $N+L$-th elements  imply the probabilities for selecting $L$ candidate numbers of denoising steps. The actual decisions on ${{x}_{n,k}}\left( t \right),{{l}_{n,k}}\left( t \right)$ can be acquired by sampling from these probabilities. Afterwards, we update ${{\Psi }_{{{n}'}}}\left( t \right),{n}'={{x}_{n,k}}\left( t \right)$ by ${{\Psi }_{{{n}'}}}\left( t \right)={{\Psi }_{{{n}'}}}\left( t \right)+{{l}_{n,k}}\left( t \right)$. According to our problem decomposition, after obtaining $\mathbf{x}\left( t \right),\mathbf{l}\left( t \right)$ from DRL actions, we derive $\mathbf{f}\left( t \right),\mathbf{b}\left( t \right)$ using the method proposed in Section IV-D.
    
    \textit{3) Reward:} Since the completion delay and energy cost of each job are affected by the total AIGC workload of the ASP executing it, the system utility $U\left( t \right)$ cannot be evaluated until the job scheduling within slot $t$ is fully completed. This results in a sparse environmental reward of the form 
    \begin{align}
    	r_{n,k}^{\text{E}}\left( t \right)\!=\!
    	\left\{\begin{array}{cl}
    		&\!\!\!\!\!\!\!\!\! U\left( t \right),\text{ if }{{j}_{n,k}}\left( t \right)\text{ is the last job in }\mathcal{J}\left( t \right),\\
    		&\!\!\!\!\!\!\!\!\!0,\qquad\text{otherwise}.
    	\end{array}\right.  \label{eq:27}
    \end{align}
    Such sparsity is further exacerbated by the term $U_{n,k}^{\text{R}}\left( t \right){{\mathbb{I}}_{\left\{ {{\Delta }_{n,k}}\left( t \right)\le {{d}_{n,k}}\left( t \right) \right\}}}$ in $U\left( t \right)$, which implies that the content generation quality of a job becomes observable only if the job is successfully completed before its deadline, otherwise the data center forcibly terminates the job to release resources. During the early stages of training, the agent is generally unable to make proper scheduling decisions to satisfy deadline constraints, resulting in low-quality and unreliable feedback. Consequently, conventional DRL algorithms struggle to make effective progress under the sparse reward signals in (\ref{eq:27}).
    
    \subsection{Diffusion Model-Aided Reward Shaping}
    
\begin{figure}[t] \centering
    	\begin{center}
    		\centerline{\includegraphics[width=7.8cm]{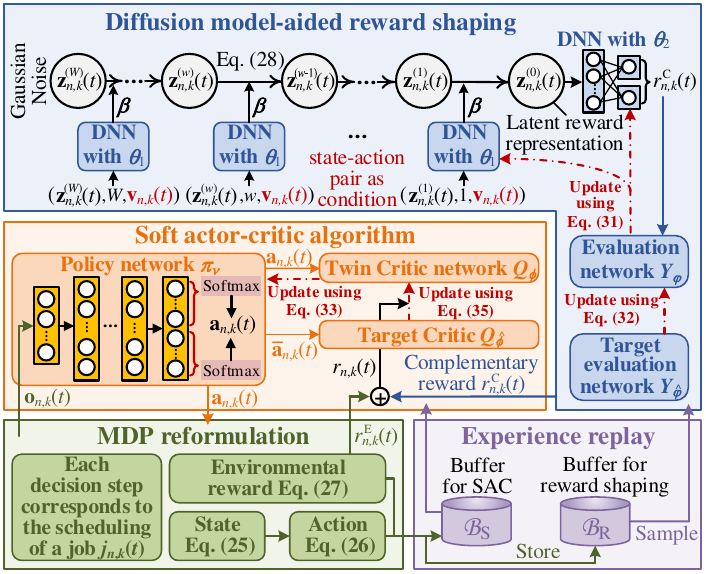}}
    	\end{center}
    	\vspace{-4mm}		
    	\caption{Diffusion model-aided reward shaping and its integration with soft actor-critic.}
    	\label{fig:3}     
    \end{figure}

    To address the intrinsic challenge of reward sparsity in AIGC workload scheduling, we propose a novel reward shaping approach that leverages the generative capability of diffusion models to synthesize complementary rewards, as illustrated in Fig. \ref{fig:3}. The key innovation lies in customizing the state-action pair ${{\mathbf{v}}_{n,k}}\left( t \right)=\left( {{\mathbf{o}}_{n,k}}\left( t \right),{{\mathbf{a}}_{n,k}}\left( t \right) \right)$ as conditioning information to guide the denoising process, thereby generating a complementary reward $r_{n,k}^{\text{C}}\left( t \right)$ that captures latent reward patterns not reflected in $r_{n,k}^{\text{E}}\left( t \right)$. We also design a tailored training mechanism for the diffusion model without requiring label data. 
    
    Specifically, the generation of $r_{n,k}^{\text{C}}\left( t \right)$ is based on the reverse diffusion process, in which an initial Gaussian noise $\mathbf{z}_{n,k}^{\left( W \right)}\left( t \right)\sim \mathcal{N}\left( \mathbf{0},{{\mathbf{I}}_{M}} \right)$ is gradually denoised for $W$ steps to yield $\mathbf{z}_{n,k}^{\left( 0 \right)}\left( t \right)$, where $M$ denotes the dimension of latent reward representation. The relationship between adjacent steps $w$ to $w-1$ is characterized by \cite{33}
	\begin{align}
		\mathbf{z}_{n,k}^{\left( w-1 \right)}\left( t \right)=&\frac{\sqrt{{{\Pi }_{w}}}\left( 1-{{{\bar{\Pi }}}_{w-1}} \right)}{1-{{\Pi }_{w}}}\mathbf{z}_{n,k}^{\left( w \right)}\left( t \right)+\frac{\sqrt{{{{\bar{\Pi }}}_{w-1}}}{{\Phi }_{w}}}{1-{{{\bar{\Pi }}}_{w}}}\nonumber\\
		&\times{{\bm{\beta} }_{{{\bm{\theta} }_{1}}}}\left( \mathbf{z}_{n,k}^{\left( w \right)}\left( t \right),w,{{\mathbf{v}}_{n,k}}\left( t \right) \right)+\sqrt{{{{\bar{\Phi }}}_{w}}}\bm{\epsilon} \label{eq:28}
	\end{align}
    where ${{\Phi }_{w}}$ is a predetermined diffusion rate at step $w$, ${{\Pi }_{w}}=1-{{\Phi }_{w}}$, ${{\bar{\Pi }}_{w}}=\prod\nolimits_{{w}'=1}^{w}{{{\Pi }_{{{w}'}}}}$, ${{\bar{\Phi }}_{w}}=\frac{1-{{{\bar{\Pi }}}_{w-1}}}{1-{{{\bar{\Pi }}}_{w}}}{{\Phi }_{w}}$ can be directly calculated. ${{\bm{\beta} }_{{{\bm{\theta} }_{1}}}}\left( \mathbf{z}_{n,k}^{\left( w \right)}\left( t \right),w,{{\mathbf{v}}_{n,k}}\left( t \right) \right)$ represents a deep neural network (DNN) parameterized by ${{\bm{\theta} }_{1}}$ that predicts the optimal latent reward representation in each denoising step $w$, conditioned on state-action pair ${{\mathbf{v}}_{n,k}}\left( t \right)$. $\bm{\epsilon}$ is randomly sampled from $\mathcal{N}\left( \mathbf{0},{{\mathbf{I}}_{L}} \right)$. After acquiring $\mathbf{z}_{n,k}^{\left( 0 \right)}\left( t \right)$, we transform it into the complementary reward $r_{n,k}^{\text{C}}\left( t \right)$ using another DNN parameterized by ${{\bm{\theta} }_{2}}$, which has two output heads that yield the mean $\mu _{n,k}^{\text{C}}\left( t \right)$ and standard deviation $\sigma _{n,k}^{\text{C}}\left( t \right)$ of $r_{n,k}^{\text{C}}\left( t \right)$, respectively. Then $r_{n,k}^{\text{C}}\left( t \right)$ is recovered by the reparameterization trick to promote exploration, i.e., 
    \begin{align}
    	r_{n,k}^{\text{C}}\left( t \right)=\tanh \left[ \mu _{n,k}^{\text{C}}\left( t \right)+\sigma _{n,k}^{\text{C}}\left( t \right)\odot \bm{\epsilon}  \right]\cdot {{e}_{1}}+{{e}_{2}}, \label{eq:29}
    \end{align}
    where $\tanh$ is used to restrict the output value for stabilizing training, $\odot$ denotes the Hadamard product, ${{e}_{1}}$ and ${{e}_{2}}$ scale the complementary reward to appropriate range. Thereafter, $r_{n,k}^{\text{C}}\left( t \right)$ is combined with the environmental reward $r_{n,k}^{\text{E}}\left( t \right)$ to form a total reward as follows
    \begin{align}
        {{r}_{n,k}}\left( t \right)=r_{n,k}^{\text{E}}\left( t \right)+\eta \cdot r_{n,k}^{\text{C}}\left( t \right), \label{eq:30}
    \end{align}
    where $\eta$ is a hyperparameter that control the weight of the complementary reward. In particular, ${{r}_{n,k}}\left( t \right)$ is received by DRL to train the agent. 
    
    Our subsequent goal is to train the DNNs with parameters $\bm{\theta} =\left\{ {{\bm{\theta} }_{1}},{{\bm{\theta} }_{2}} \right\}$ used for reward shaping. Since the label for $r_{n,k}^{\text{C}}\left( t \right)$ is difficult to collect, we introduce an evaluation network ${{Y}_{\bm{\varphi} }}( {{\mathbf{v}}_{n,k}}\left( t \right),r_{n,k}^{\text{C}}\left( t \right) )$ parameterized by $\bm{\varphi}$ to estimate the contribution of the complementary reward to improving long-term accumulative return. Therefore, the training loss for $\bm{\theta}$ is expressed as
    \begin{align}
    	L\left( \bm{\theta}  \right)=-\mathbb{E}\left[ {{Y}_{\bm{\varphi} }}\left( {{\mathbf{v}}_{n,k}}\left( t \right),r_{n,k}^{\text{C}}\left( t \right) \right) \right]. \label{eq:31}
    \end{align}
    To update $\bm{\varphi} $, a target evaluation network ${{Y}_{{\hat{\bm{\varphi} }}}}$ is invoked, whose parameter $\hat{\bm{\varphi} }$ is updated via tardily following $\bm{\varphi} $. Then the training loss for $\bm{\varphi} $ is
    \begin{align}
    	L\left( \bm{\varphi}  \right)=\mathbb{E}[ r_{n,k}^{\text{E}}\left( t \right)&+\Gamma {{Y}_{{\hat{\bm{\varphi} }}}}\left( {{{\mathbf{\bar{v}}}}_{n,k}}\left( t \right),\bar{r}_{n,k}^{\text{C}}\left( t \right) \right)\nonumber\\
    	&\qquad-{{Y}_{\bm{\varphi} }}\left( {{\mathbf{v}}_{n,k}}\left( t \right),r_{n,k}^{\text{C}}\left( t \right) \right) ]^{2}, \label{eq:32}
    \end{align}
    where $\Gamma$ is the discount factor, ${{\mathbf{\bar{v}}}_{n,k}}\left( t \right)=\left( {{{\mathbf{\bar{o}}}}_{n,k}}\left( t \right),{{{\mathbf{\bar{a}}}}_{n,k}}\left( t \right) \right)$ indicates the state-action pair of the next environmental step, and $\bar{r}_{n,k}^{\text{C}}\left( t \right)$ is generated conditioned on ${{\mathbf{\bar{v}}}_{n,k}}\left( t \right)$. 
    
    \subsection{Integrating Reward Shaping With Soft Actor-Critic}
     
    It is worth noting that the proposed diffusion model-aided reward shaping serves as a general approach compatible with various DRL paradigms. In this work, we integrate it with soft actor-critic (SAC) to take advantage of its maximum-entropy objective and off-policy learning capability \cite{34}. SAC adopts a policy network ${{\pi }_{\bm{\nu} }}$ parameterized by $\bm{\nu} $ to output action ${{\mathbf{a}}_{n,k}}\left( t \right)$ based on state ${{\mathbf{o}}_{n,k}}\left( t \right)$. In order to satisfy constraints (\ref{eq:21b})-(\ref{eq:21d}), we adapt the output layer of ${{\pi }_{\bm{\nu} }}$ to generate a $N+L$-dimensional vector, then apply the softmax activation function to attain the probabilities in (\ref{eq:26}). If the number of transferred jobs from ASP $n$ is already equal to $J_{n}^{\max }$, we force ${{x}_{n,k}}\left( t \right)$ to be $n$. Additionally, twin critic networks ${{Q}_{{{\bm{\phi} }_{1}}}},{{Q}_{{{\bm{\phi} }_{2}}}}$ are considered to evaluate soft Q values based on the state-action pair while mitigating overestimation. The corresponding target networks ${{Q}_{{{{\hat{\bm{\phi} }}}_{1}}}},{{Q}_{{{{\hat{\bm{\phi} }}}_{2}}}}$ are used to prevent training oscillation. 
    
	The training loss for the policy network is derived from the Kullback-Leibler (KL) divergence between policy and Q value distribution, written as
	\begin{align}
		\!\! L\!\left( \bm{\nu}  \right)\!=\!\mathbb{E}\big[ \Xi \ln {{\pi }_{\bm{\nu} }}\left( {{\mathbf{a}}_{n,k}}\!\left( t \right)\!|{{\mathbf{o}}_{n,k}}\!\left( t \right) \right)\!-\!\!\underset{i=1,2}{\mathop{\min }}\,{{Q}_{{{\bm{\phi} }_{i}}}}\!\!\left( {{\mathbf{v}}_{n,k}}\!\left( t \right) \right) \big]\!,\!\!\label{eq:33}
	\end{align}
	where $\Xi$ denotes the temperature parameter, which is automatically adjusted by minimizing
	\begin{align}
		L\left( \Xi  \right)=\mathbb{E}\left[ -\Xi \ln {{\pi }_{\bm{\nu} }}\left( {{\mathbf{a}}_{n,k}}\left( t \right)|{{\mathbf{o}}_{n,k}}\left( t \right) \right)+\left( N+L \right)\Xi  \right], \label{eq:34}
	\end{align}
	The critic loss is given by
	\begin{align}
		&\! L\left( {{\bm{\phi} }_{i}} \right)=\mathbb{E}\big[ {{r}_{n,k}}\left( t \right)+\Gamma \underset{i=1,2}{\mathop{\min }}\,{{Q}_{{{{\hat{\bm{\phi} }}}_{i}}}}\left( {{{\mathbf{\bar{v}}}}_{n,k}}\left( t \right) \right)\nonumber\\
		&\! -\!\Xi \ln\!{{\pi }_{\bm{\nu} }}\!\left( {{{\mathbf{\bar{a}}}}_{n,k}}\!\left( t \right)\!|{{{\mathbf{\bar{o}}}}_{n,k}}\!\left( t \right) \right)\!-\!{{Q}_{{{\bm{\phi} }_{i}}}}\!\left( {{\mathbf{v}}_{n,k}}\!\left( t \right) \right) \!\big]^{2},i\!\in\!\left\{ 1,2 \right\}. \label{eq:35}
	\end{align}
    
    During training, we maintain two separate replay buffers, ${{\mathcal{B}}_{\text{S}}}$ and ${{\mathcal{B}}_{\text{R}}}$, for the SAC agent and reward shaping model, respectively. After each interaction with the environment, experience tuple in the form $\left( {{\mathbf{o}}_{n,k}}\left( t \right),{{\mathbf{a}}_{n,k}}\left( t \right),r_{n,k}^{\text{E}}\left( t \right),{{{\mathbf{\bar{o}}}}_{n,k}}\left( t \right) \right)$ is stored in ${{\mathcal{B}}_{\text{S}}}$. The diffusion model is then invoked to generate a complementary reward, after which $\left( {{\mathbf{v}}_{n,k}}\left( t \right),r_{n,k}^{\text{C}}\left( t \right),r_{n,k}^{\text{E}}\left( t \right),{{{\mathbf{\bar{v}}}}_{n,k}}\left( t \right) \right)$ is stored in ${{\mathcal{B}}_{\text{R}}}$. All DNNs are trained in an off-policy manner, i.e., by sampling mini-batches from ${{\mathcal{B}}_{\text{S}}}$ or ${{\mathcal{B}}_{\text{R}}}$ to compute the loss functions. An implementation detail worth emphasizing is that $r_{n,k}^{\text{C}}\left( t \right)$ used in (\ref{eq:31}) and (\ref{eq:35}) is generated by the most recently updated diffusion model. 
	
	\subsection{GPU DVFS and Power Usage Behavior Optimization}
	
	In this part, we delve into subproblem \textbf{SP2}$_{n}\left( t \right)$ in (\ref{eq:24}). To deal with the intractable indicator function ${{\mathbb{I}}_{\left\{ {{\Delta }_{k}}\left( t \right)\le {{d}_{k}}\left( t \right) \right\}}}$, an auxiliary variable ${{\alpha }_{k}}\left( t \right)\in \left\{ 0,1 \right\}$ is introduced to indicate whether job $k$ can be completed in time, i.e., ${{\alpha }_{k}}\left( t \right)=1$ if ${{\Delta }_{k}}\left( t \right)\le {{d}_{k}}\left( t \right)$, otherwise ${{\alpha }_{k}}\left( t \right)=0$. Denote $\bm{\alpha} \left( t \right)=\left\{ {{\alpha }_{k}}\left( t \right):\forall k \right\}$, then (\ref{eq:24}) can be recast as
	\begin{subequations}
		\begin{align}
				\textbf{SP2.1}_n(t): &\underset{{{\mathbf{f}}_{n}}\left( t \right),{{\mathbf{b}}_{n}}\left( t \right),\bm{\alpha} \left( t \right)}{\mathop{\max }}\sum\limits_{k\in \mathcal{K}_{n}^{*}\left( t \right)}{{{\alpha }_{k}}\left( t \right)\left[ U_{k}^{\text{R}}\left( t \right)+\upsilon  \right]}\nonumber\\
				&\qquad-{{\varsigma }_{n}}\left( t \right){{G}_{n}}\left( t \right)\tau -\Upsilon {{\tilde{E}}_{n}}\left( t \right){{D}_{n}}\left( t \right)\tau, \label{eq:36a}
		\end{align}
		\vspace{-5mm}
		\begin{align}
			\mbox{s.t.}\ 
			&{{\Delta }_{k}}\left( t \right){{\alpha }_{k}}\left( t \right)\le {{d}_{k}}\left( t \right),\ \forall k, \label{eq:36b}\\
			&\text{(\ref{eq:23}), (\ref{eq:21e})-(\ref{eq:21h}), and (\ref{eq:9})-(\ref{eq:10}) evaluated at }(n,t), \nonumber
		\end{align} \label{eq:36}
	\end{subequations}
	$\!\!$which remains a challenging MINLP problem. The computational complexity is prohibitive for enumerating all possible $\bm{\alpha} \left( t \right)$ when job scale grows. To efficiently address \textbf{SP2.1}$_{n}\left( t \right)$, we propose a greedy heuristic method to iteratively find the near-optimal $\bm{\alpha} \left( t \right)$, and derive closed-form expressions of ${{\mathbf{f}}_{n}}\left( t \right),{{\mathbf{b}}_{n}}\left( t \right)$ for each given $\bm{\alpha} \left( t \right)$. 
	
	Specifically, based on ${{\bm{\alpha} }^{i-1}}\left( t \right)$ from the previous iteration $i-1$, our greedy method evaluates the incremental objective value associated with each candidate ${{\alpha }_{k}}\left( t \right)$ and identify ${{k}^{*}}$ that yields the maximum objective improvement, while ensuring that the delay constraint is satisfied. We then set ${{\alpha }_{{{k}^{*}}}}\left( t \right)=1$ to update ${{\bm{\alpha} }^{i}}\left( t \right)$, and the iterative process continues until either all ${{\alpha }_{k}}\left( t \right)$ are set to 1 or the delay constraint cannot be satisfied by the remaining jobs. 
	
	Subsequently, we provide the closed-form solutions of ${{\mathbf{f}}_{n}}\left( t \right),{{\mathbf{b}}_{n}}\left( t \right)$ as well as feasibility assessment criteria in the following theorem. 
	
	\textit{Theorem 1:} Given any job completion indicator $\bm{\alpha} \left( t \right)$, the optimal GPU DVFS is given by
	\begin{align}
		&f_{n}^{\text{c*}}\left( t \right)=
		\left\{\begin{array}{cl}
			&\!\!\!\!\!\!\!\!\! f_{n}^{\text{c},\min },\ \tilde{f}_{n}^{\text{c}}\left( t \right)\le f_{n}^{\text{c},\min },\\
			&\!\!\!\!\!\!\!\!\!\tilde{f}_{n}^{\text{c}}\left( t \right),\ f_{n}^{\text{c},\min }<\tilde{f}_{n}^{\text{c}}\left( t \right)\le h\left[ V_{n}^{\text{c}}\left( t \right) \right],\\
			&\!\!\!\!\!\!\!\!\! h\left[ V_{n}^{\text{c}}\left( t \right) \right],\ \tilde{f}_{n}^{\text{c}}\left( t \right)>h\left[ V_{n}^{\text{c}}\left( t \right) \right],
		\end{array}\right.  \label{eq:37}\\
	    &f_{n}^{\text{m*}}\left( t \right)=\max \left\{ \frac{{{\varepsilon }_{n}}\left( 1-{{\kappa }_{n}} \right)}{{{{\tilde{d}}}^{\min }}\left( t \right)-\Delta _{n}^{0}-\frac{{{\varepsilon }_{n}}{{\kappa }_{n}}}{f_{n}^{\text{c*}}\left( t \right)}},f_{n}^{\text{m},\min } \right\}, \label{eq:38}
	\end{align}
    where $\tilde{f}_{n}^{\text{c}}\left( t \right)=\frac{{{\varepsilon }_{n}}{{\kappa }_{n}}}{{{{\tilde{d}}}^{\min }}\left( t \right)-\Delta _{n}^{0}}+\frac{{{\varepsilon }_{n}}\sqrt{{{{\lambda }_{n}}\left( 1-{{\kappa }_{n}} \right){{\kappa }_{n}}}/{{{\delta }_{n}}}\;}}{\left[ {{{\tilde{d}}}^{\min }}\left( t \right)-\Delta _{n}^{0} \right]V_{n}^{\text{c}}\left( t \right)}$, ${{\tilde{d}}^{\min }}\left( t \right)$ is the minimum delay requirement among all ${{\tilde{d}}_{k}}\left( t \right)$ with ${{\alpha }_{k}}\left( t \right)=1$, and ${{\tilde{d}}_{k}}\left( t \right)=\frac{2{{\Theta }_{n}}\left[ {{d}_{k}}\left( t \right)-\Delta _{k}^{\text{tra}}\left( t \right)-\Delta _{k}^{\text{fee}}\left( t \right) \right]}{\sum\nolimits_{{k}'\in \mathcal{K}_{n}^{*}\left( t \right)}{{{l}_{{{k}'}}}\left( t \right)}+{{l}_{k}}\left( t \right)}$. The optimal $V_{n}^{\text{c*}}\left( t \right)$ can be found via a simply one-dimensional search over $\left[ V_{n}^{\text{c},\min },V_{n}^{\text{c},\max } \right]$, which combines (\ref{eq:37})-(\ref{eq:38}) to achieve the minimum GPU power consumption $P_{n}^{\text{G*}}\left( t \right)$. Moreover, the optimal cooling and discharging power are determined by
	\begin{align}
		&P_{n}^{\text{C*}}\!\left( t \right)\!=\!-\min \left\{ \frac{\tilde{\zeta }_{n}^{\max }\left( t \right)}{{{\vartheta }^{\text{COP}}}},\frac{\Delta \tilde{\zeta }_{n}^{\max }\left( t \right)}{{{\vartheta }^{\text{COP}}}} \right\}\!+\!\frac{P_{n}^{\text{G*}}\left( t \right)}{{{\vartheta }^{\text{COP}}}}, \label{eq:39}\\
		&D_{n}^{*}\!\left( t \right)\!=\!
		\left\{\begin{array}{cl}
			&\!\!\!\!\!\!\!\!\! -\!\min \left\{ D_{n}^{\text{min}},{{{{\tilde{E}}}_{n}}\left( t \right)}/{\tau } \right\}\!,{{\varsigma }_{n}}\left( t \right)\!\le\! \Upsilon {{{\tilde{E}}}_{n}}\left( t \right), \\
			&\!\!\!\!\!\!\!\!\! \min \left\{ D_{n}^{\text{max}},{{{E}_{n}}\left( t \right)}/{\tau } \right\}\!,\ {{\varsigma }_{n}}\left( t \right)\!>\!\Upsilon {{{\tilde{E}}}_{n}}\left( t \right).
		\end{array}\right.  \label{eq:40}
	\end{align}
	Regarding feasibility assessment, if there is no $V_{n}^{\text{c}}\left( t \right)$ such that $f_{n}^{\text{m*}}\left( t \right)\le f_{n}^{\text{m},\max }$, or the calculated $P_{n}^{\text{C*}}\left( t \right)>P_{n}^{\text{C},\max }$, the problem is infeasible under the current $\bm{\alpha} \left( t \right)$.
	
	\textit{Proof:} Please refer to Appendix A. $\qquad\qquad\qquad\qquad\ \blacksquare$
    
    \subsection{Overall Solution Algorithm}
    
    \begin{algorithm}[t]
    	\caption{Joint Energy Management and AIGC Workload Scheduling Algorithm (JEMAS)} \label{alg:1}
    	\begin{algorithmic}[1]
    		\STATE \textbf{Initialize:} Parameters of reward shaping model $\bm{\theta} ,\bm{\varphi} ,\hat{\bm{\varphi} }$, SAC agent $\bm{\nu} ,{{\bm{\phi} }_{1}},{{\bm{\phi} }_{2}},{{\hat{\bm{\phi} }}_{1}},{{\hat{\bm{\phi} }}_{2}}$, temperature parameter $\Xi$, replay buffers ${{\mathcal{B}}_{\text{S}}},{{\mathcal{B}}_{\text{R}}}$.
    		\FOR{each training episode} 
    		\FOR{$t\in \mathcal{T}$}
    		\FOR{${{j}_{n,k}}\left( t \right)\in \mathcal{J}\left( t \right)$}
    		\STATE Take action ${{\mathbf{a}}_{n,k}}\left( t \right)$ using policy ${{\pi }_{\bm{\nu} }}\left( \cdot |{{\mathbf{o}}_{n,k}}\left( t \right) \right)$.
    		\IF{${{j}_{n,k}}\left( t \right)$ is the last job in $\mathcal{J}\left( t \right)$}
    		\STATE Obtain $\mathbf{x}\left( t \right),\mathbf{l}\left( t \right)$ from DRL actions.
    		\FOR{$n\in \mathcal{N}$}
    		\STATE Optimize ${{\mathbf{f}}_{n}}\left( t \right),{{\mathbf{b}}_{n}}\left( t \right)$ using greedy heuristic and closed-form solutions in (\ref{eq:37})-(\ref{eq:40}).
    		\ENDFOR
    		\ENDIF
    		\STATE Calculate environmental reward $r_{n,k}^{\text{E}}\left( t \right)$ and observe next state ${{\mathbf{\bar{o}}}_{n,k}}\left( t \right)$.
    		\STATE Generate complementary reward $r_{n,k}^{\text{C}}\left( t \right)$ using diffusion process (\ref{eq:28}) and reparameterization (\ref{eq:29}).
    		\STATE Store $( {{\mathbf{o}}_{n,k}}\left( t \right),{{\mathbf{a}}_{n,k}}\left( t \right),r_{n,k}^{\text{E}}\left( t \right),{{{\mathbf{\bar{o}}}}_{n,k}}\left( t \right) )$ in ${{\mathcal{B}}_{\text{S}}}$.
    		\STATE Store $( {{\mathbf{v}}_{n,k}}\left( t \right),r_{n,k}^{\text{C}}\left( t \right),r_{n,k}^{\text{E}}\left( t \right),{{{\mathbf{\bar{v}}}}_{n,k}}\left( t \right) )$ in ${{\mathcal{B}}_{\text{R}}}$. 
    		\ENDFOR
    		\STATE Sample a mini-batch of tuples from ${{\mathcal{B}}_{\text{S}}}$.
    		\STATE Update policy network ${{\pi }_{\bm{\nu} }}$ by minimizing (\ref{eq:33}). 
    		\STATE Compute total reward ${{r}_{n,k}}\left( t \right)$ using (\ref{eq:30}).
    		\STATE Update critic networks $\!{{Q}\!_{{{\bm{\phi} }_{1}}}}\!,\!{{Q}\!_{{{\bm{\phi} }_{2}}}}\!$ by minimizing (\ref{eq:35}).
    		\STATE Update $\Xi$ by minimizing (\ref{eq:34}).
    		\STATE Sample a mini-batch of tuples from ${{\mathcal{B}}_{\text{R}}}$.
    		\STATE Update $\bm{\theta}$ for reward shaping by minimizing (\ref{eq:31}).
    		\STATE Update evaluation network ${{Y}_{\bm{\varphi} }}$ by minimizing (\ref{eq:32}).
    		\STATE Slowly update all target networks. 
    		\ENDFOR
    		\ENDFOR
    	\end{algorithmic}
    \end{algorithm}

	Based on the above elaborations, the developed \underline{J}oint \underline{E}nergy \underline{M}anagement and \underline{A}IGC workload \underline{S}cheduling (JEMAS) algorithm is outlined in \textbf{Algorithm 1}. In each time slot, the agent makes job scheduling decisions using the policy network, then the greedy heuristic is employed to optimize GPU DVFS and power usage behavior. Upon receiving the environmental reward, a complementary reward is generated via the diffusion process, forming experience tuples for replay. Subsequently, we update both the SAC agent and the reward shaping model in an off-policy manner. This procedure repeats until the maximum number of training episodes is reached. During implementation, only the SAC agent undergoes forward propagation, with lightweight metadata like renewable generation and electricity price exchanged among ASPs to construct the state. The diffusion model incurs no additional computational overhead, thereby ensuring the timeliness of decision-making. The algorithm optimality is theoretically ensured as follows. 
	
	\textit{Theorem 2:} Under the assumption that the evaluation network ${{Y}_{\bm{\varphi} }}$ is well-trained with $L\left( \bm{\varphi}  \right)=0$, the complementary reward generated by the diffusion model is consistent with the potential-based shaping structure in \cite{37}, which preserves the optimality of the original system utility maximization problem.
	
	\textit{Proof:} Please refer to Appendix B. $\qquad\qquad\qquad\qquad\ \blacksquare$
    
	\section{Performance Evaluation} \label{sec:evaluation}
	
	\subsection{Simulation Setting}
	
	In this section, we conduct simulations to evaluate the performance of the proposed JEMAS algorithm. The default parameter settings are summarized in Table \ref{tab:I}. The job arrival patterns of $N=3$ ASPs are generated based on distinct real-world traces from Alibaba, as illustrated in Fig. \ref{fig:4} (a), thereby capturing practical traffic diversity across ASPs\footnote{https://github.com/alibaba/clusterdata/tree/master}. All jobs are considered to be transferable with $\left\{ {{\gamma }_{n,k}}\left( t \right)=1:\forall n,t \right\}$ to draw fundamental insights. The set of denoising steps is $\mathcal{L}=\left\{ 6,10,14,\ldots,38,42 \right\}$. Job prompts are sampled from the PartiPrompts dataset\footnote{https://github.com/google-research/parti/blob/main/PartiPrompts.tsv}. To capture the variability of renewable energy generation, we assume that solar panels and wind turbines with capacities of 10 MW and 22.4 MW are deployed at ASP 2 and ASP 3, respectively, with real-world output profiles\footnote{https://www.nrel.gov/grid/solar-integration-data}. The electricity prices are adopted from the PJM market data, different ASPs experience distinct price profiles\footnote{https://www.pjm.com/markets-and-operations}, as shown in Fig. \ref{fig:4} (b)-(d). 
	
	\begin{table}[t]\footnotesize
		\centering
		\setlength{\abovecaptionskip}{0pt}    
		\setlength{\belowcaptionskip}{10pt}
		\caption{Simulation Parameters} \label{tab:I}
		\begin{threeparttable}
			\begin{tabular}{c|c|c|c}
				\hline
				\textbf{Parameter}&\textbf{Value}&\textbf{Parameter}&\textbf{Value}\\
				\hline
				$N$&3&$T$&288\\
				$\tau$&5 min&Job scale&3638\\
				${{d}_{n,k}}\left( t \right)$&[30, 270] s&${{\vartheta }^{\text{COP}}},P_{n}^{\text{C},\max }$&3, 10 MW\\
				$\Upsilon$&8&$\rho c{{V}_{n}}$&0.0016 MWh/$^{\circ}$C\\
				$\Delta \zeta _{n}^{\text{in},\max }$&2 $^{\circ}$C&$\zeta _{n}^{\text{in},\max },\zeta _{n}^{\text{in},\max }$&30 $^{\circ}$C, 34 $^{\circ}$C\\
				$E_{n}^{\max }$&15 MWh&$D_{n}^{\text{min}},D_{n}^{\text{max}}$&3 MW, 3 MW\\
				$\upsilon$&150 &$\psi$&0.0625 \$/MB\\
				\multirow{2}{*}{$W$}&\multirow{2}{*}{3}&\multirow{2}{*}{${{\Phi }_{w}}$}&Variational posterior\\
				& & &scheduler \cite{9}\\
				$\eta $&0.4&$\Gamma$&0.99\\
				$M$&128&Learning rate&3$\times$10$^\text{-4}$\\
				\hline
			\end{tabular}
		\end{threeparttable}			
	\end{table} 
    
    For diffusion-based AIGC models, we consider that Stable Diffusionv1-5 (SD1.5), SDXL, and SD3.5\footnote{https://huggingface.co/stabilityai} are hosted at the three ASPs, respectively. Their parameter sizes range from 1.06B to 8.1B, leading to heterogeneous image generation performance and computational workloads. In the aforementioned Fig. \ref{fig:2}, we evaluate the BRISQUE and CLIP scores of the content generated by different AIGC models under varying numbers of denoising steps, where each data point is obtained by averaging over 50 images generated from 10 text prompts, each with 5 random seeds. The fitted sigmoid functions are used to compute AIGC service revenues in the following experiments. Additionally, Fig. \ref{fig:2} incorporates FLUX.1-dev, a diffusion-transformer model distinct from Stable Diffusion, demonstrating that the proposed service revenue modeling can capture heterogeneous AIGC architectures in a unified manner.
    
    Regarding GPU DVFS settings, we adopt normalized values of $\left( f_{n}^{\text{c}}\left( t \right),V_{n}^{\text{c}}\left( t \right),f_{n}^{\text{m}}\left( t \right) \right)$ without loss of generality, where (1,1,1) corresponding to (1880 MHz, 1.05 V, 6300 MHz) \cite{29}. The DVFS scaling intervals are set to be $f_{n}^{\text{c}}\left( t \right)\in [ 0.5,\sqrt{{\left[ V_{n}^{\text{c}}\left( t \right)-0.5 \right]}/{2}}+0.5 ]$, $V_{n}^{\text{c}}\left( t \right)\in \left[ 0.5,1.2 \right]$, and $f_{n}^{\text{m}}\left( t \right)\in \left[ 0.5,1.2 \right]$. Based on the measured average runtime power and execution time when running different AIGC models, we set $P_{n}^{0}\in \left\{ 3,4,5 \right\}$, ${{\lambda }_{n}}\in \left\{ 1.5,2,2.5 \right\}$, ${{\delta }_{n}}\in \left\{ 4.5,6,7.5 \right\}$ to compute $P_{n}^{\text{G}}\left( t \right)$ in (\ref{eq:2}), and $\Delta _{n}^{0}\in \left\{ 0.028,0.0667,0.1489 \right\}$, ${{\varepsilon }_{n}}\in \left\{ 0.14,0.3332,0.7448 \right\}$, ${{\kappa }_{n}}=0.5$ to compute $\Delta _{n,k}^{\text{exe}}\left( t \right)$ in (\ref{eq:3}).  
    
    The neural network architectures adopted in JEMAS are given below. The DNN with ${\bm{\theta }_{1}}$ uses sinusoidal positional embeddings and two hidden layers to project the input into a 256-dimensional representation, followed by two 256-neuron hidden layers for output generation. The DNN with ${\bm{\theta }_{2}}$ employs two output heads to map the 128-dimensional latent reward to its mean and standard deviation. Both the policy and critic networks in SAC consist of two hidden layers with 256 neurons. 
    
    \subsection{AIGC Workload Scheduling Results}
    
\begin{figure*}[t] \centering
		\subfigure[Job arrival pattern. ] {   
			{\includegraphics[width=3.3cm]{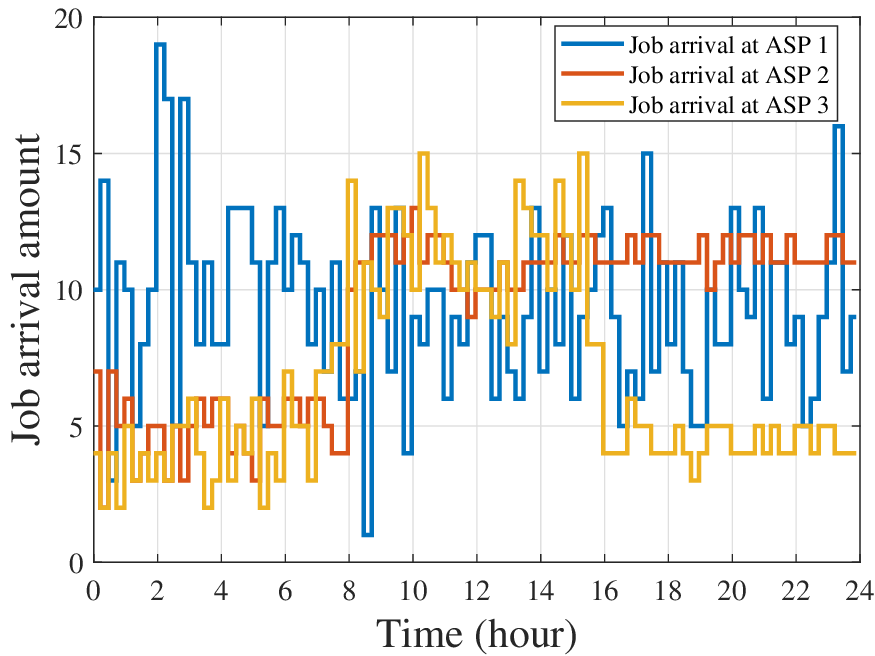}}       
		}
		\subfigure[Denoising steps executed at ASP 1.] { 
			{\includegraphics[width=4.5cm]{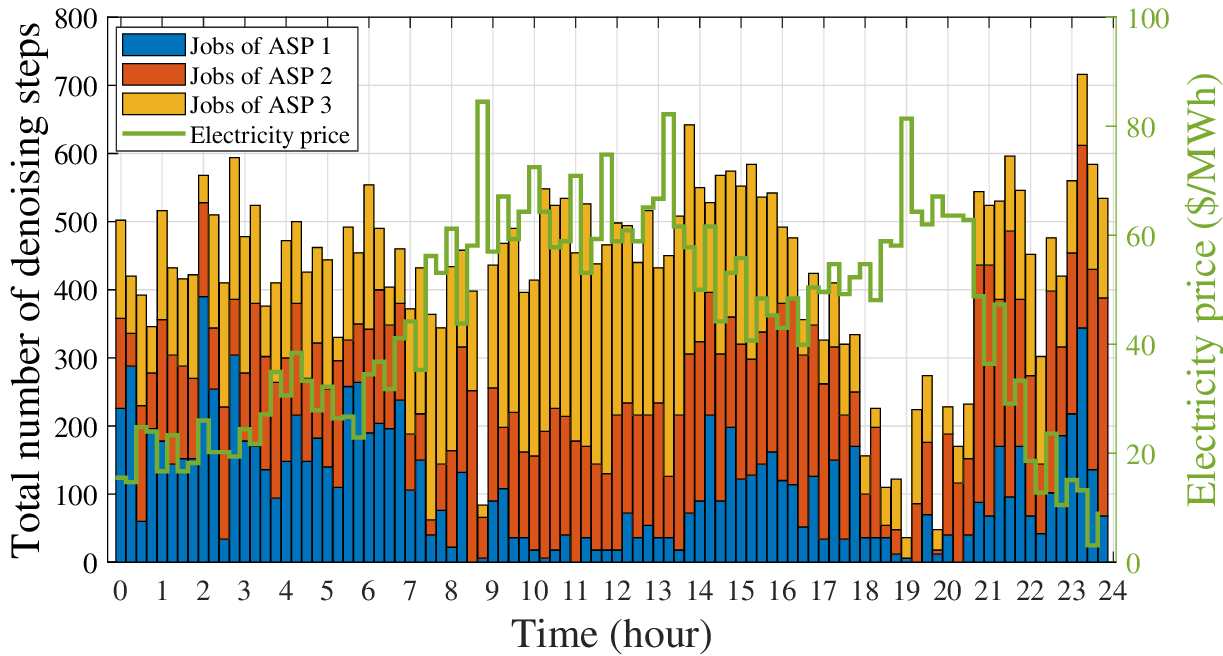}} 
		}     
		\subfigure[Denoising steps executed at ASP 2.] {   
			{\includegraphics[width=4.5cm]{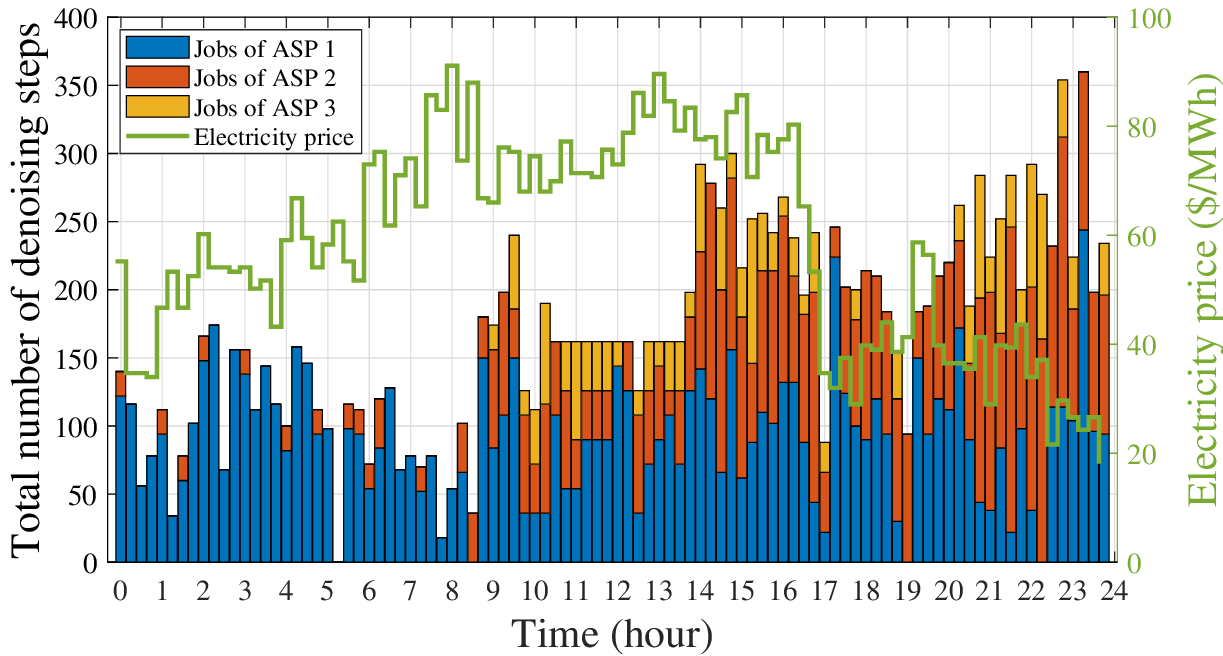}}       
		}
		\subfigure[Denoising steps executed at ASP 3. ] {   
			{\includegraphics[width=4.5cm]{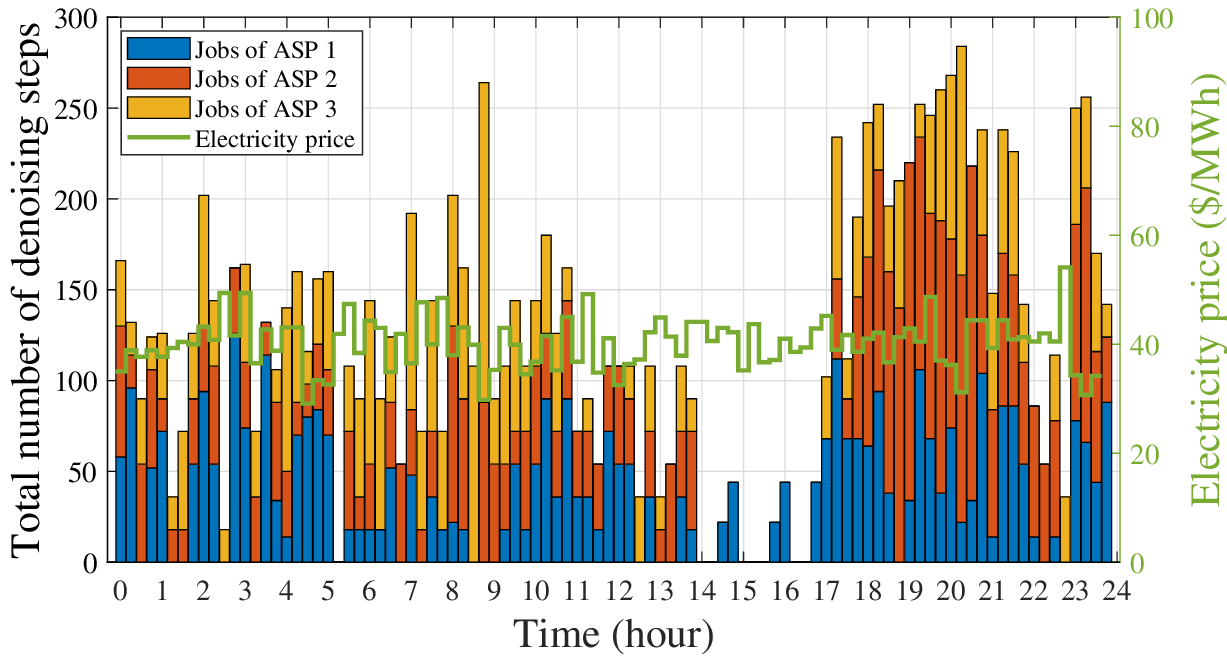}}       
		}
		\vspace{1mm}
		\caption{Job arrival and execution results of ASPs. }     
		\label{fig:4}     
	\end{figure*}

    Fig. \ref{fig:4} (a)-(c) illustrate the total numbers of denoising steps executed at the three ASPs, which reflect the temporal variations in computational workload. As observed, each ASP tends to execute fewer denoising steps during periods of high electricity prices, thereby reducing energy costs and improving system utility. For instance, the workloads of ASP 1 and ASP 2 remain at relatively low levels during 17:00-20:00 and 7:00-9:00, respectively. This behavior is attributed to both job transferring toward lower-price locations and adaptive denoising step configuration, enabling flexible responses to electricity price fluctuations.
    
    Meanwhile, workload scheduling is influenced by the AIGC models deployed at the destination ASPs. Specifically, jobs processed at ASP 1 are typically assigned larger denoising step configurations, whereas those at ASPs 2 and 3 adopt fewer steps. This is because SD1.5 deployed at ASP 1 exhibits relatively lower generation performance, requiring more denoising steps to enhance content quality. In contrast, ASP 3, equipped with SD3.5, can achieve satisfactory content quality with fewer steps, thereby reducing completion delay and energy consumption. 
	
	\begin{figure*}[t] \centering
		\subfigure[ASP 1.] { 
			{\includegraphics[width=4.4cm]{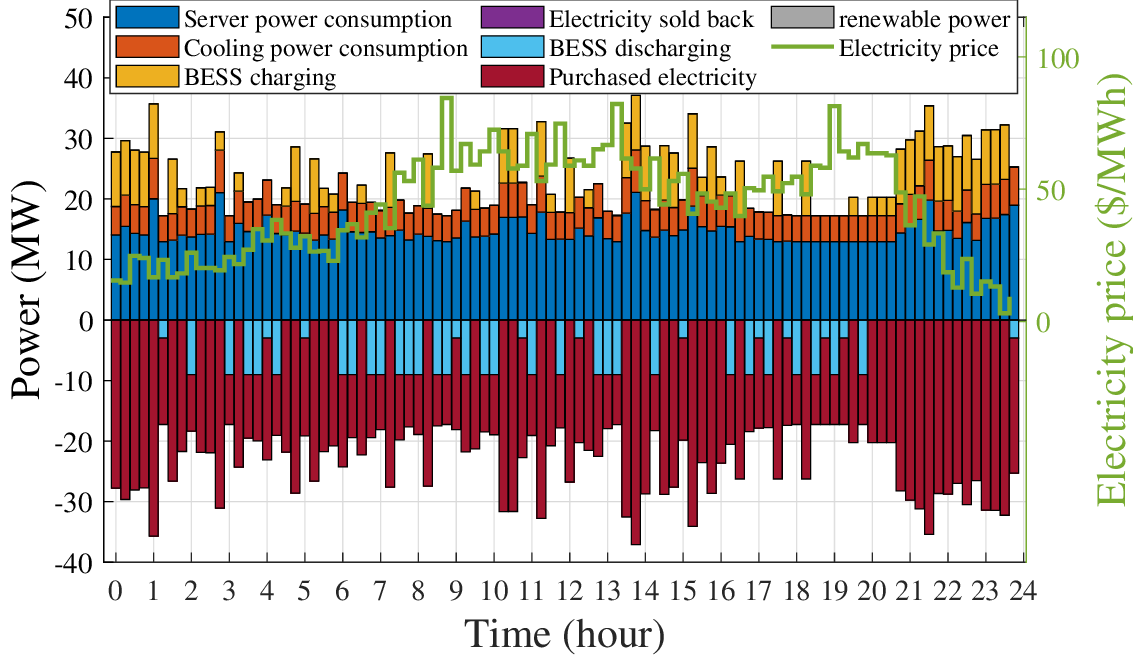}} 
		}     
		\subfigure[ASP 2.] {   
			{\includegraphics[width=4.4cm]{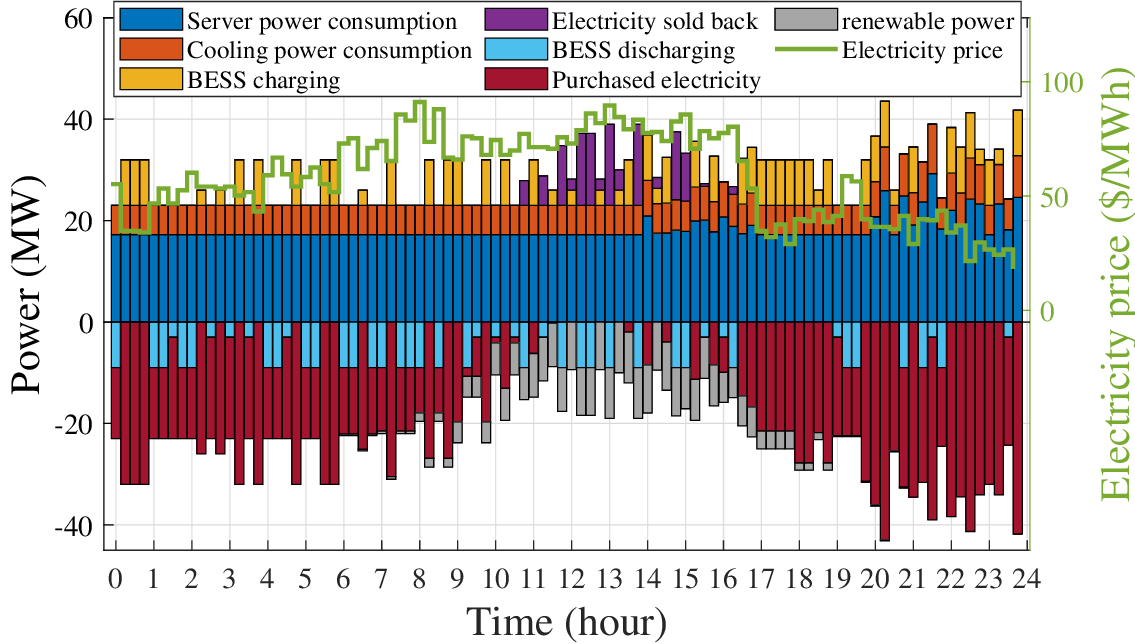}}       
		}
		\subfigure[ASP 3. ] {   
			{\includegraphics[width=4.4cm]{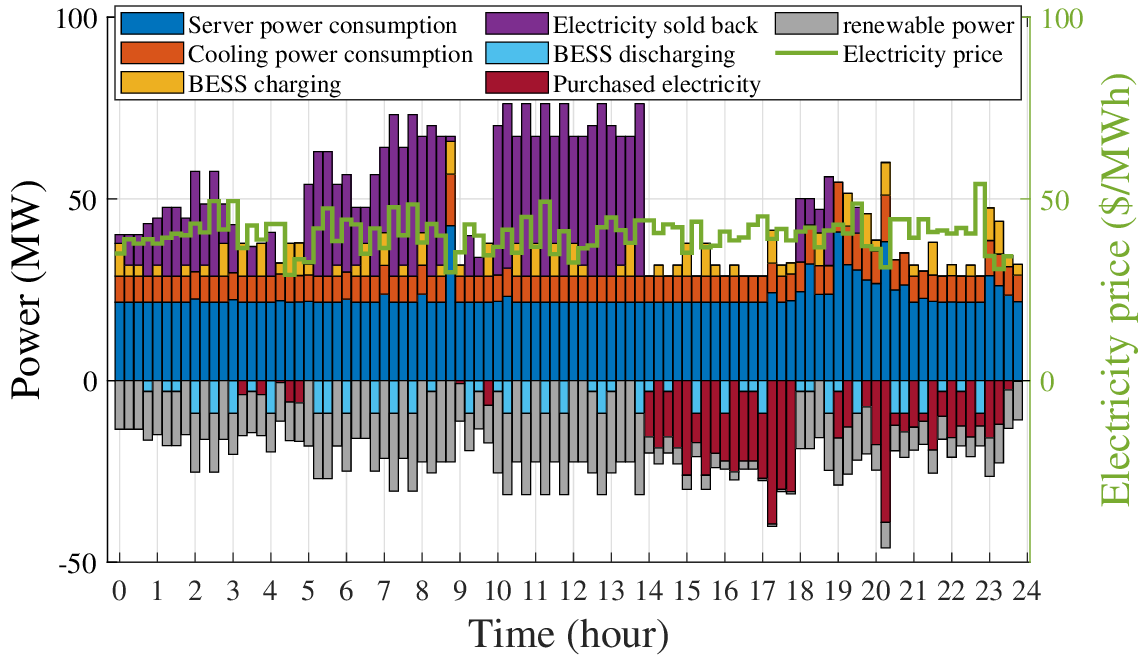}}       
		}
		\subfigure[Temperature evolution. ] {   
			{\includegraphics[width=3.5cm]{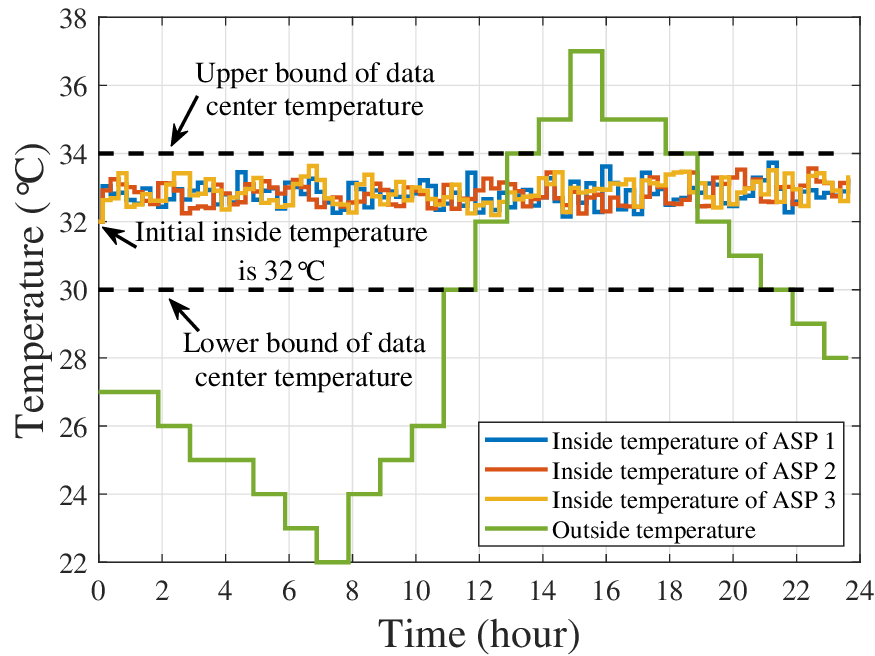}}       
		}
		\vspace{1mm}
		\caption{Energy management results of ASPs. }     
		\label{fig:5}     
	\end{figure*}
    
    Fig. \ref{fig:5} (a)-(c) present the energy management results of the three ASPs, where power consumption components are shown as positive values and power supply components as negative values. We can observe that server power consumption varies over time under adaptive GPU DVFS, with data center energy resources allocated accordingly to accommodate fluctuating computational workloads. For example, due to low electricity prices at ASP 1 during 0:00-6:00, its BESS is charged to store energy for future uncertainties. During 12:00-16:00, ASP 2 experiences high electricity prices and therefore prioritizes the utilization of renewable energy and BESS stored energy to meet power demand, thereby reducing grid purchases. At ASP 3, surplus renewable generation during 10:00-14:00 is sold back to the grid to obtain additional revenue. Fig. \ref{fig:5} (d) showcases the temperature evolution, where the data center inside temperatures of all ASPs are constantly within the allowable range. This is because the temperature constraints are actively forced during cooling power optimization. 
	
	\begin{figure}[t] \centering
		\subfigure[ASP selection.] { 
			{\includegraphics[width=4.1cm]{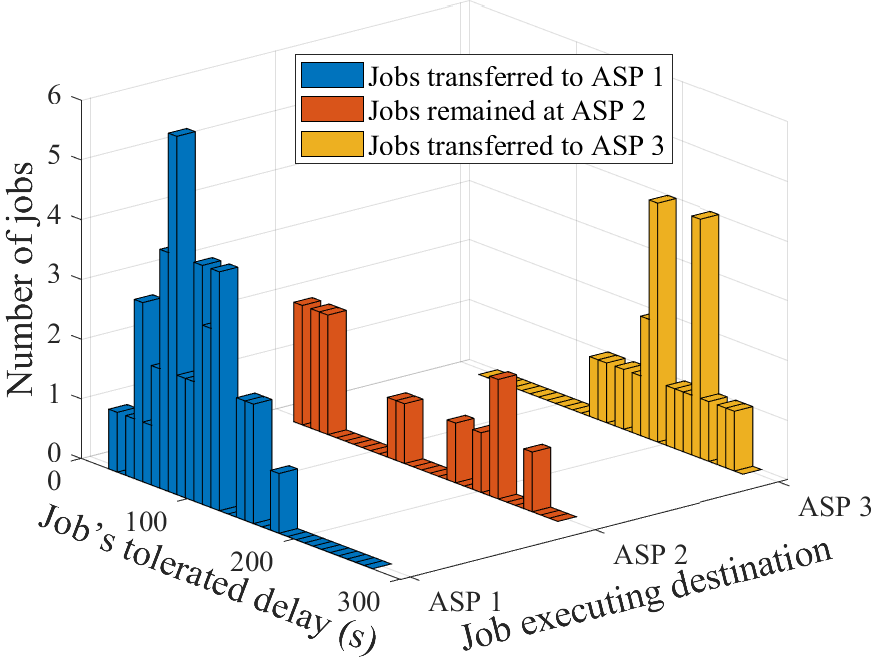}} 
		}     
		\subfigure[Denoising step configuration.] { 
			{\includegraphics[width=4.1cm]{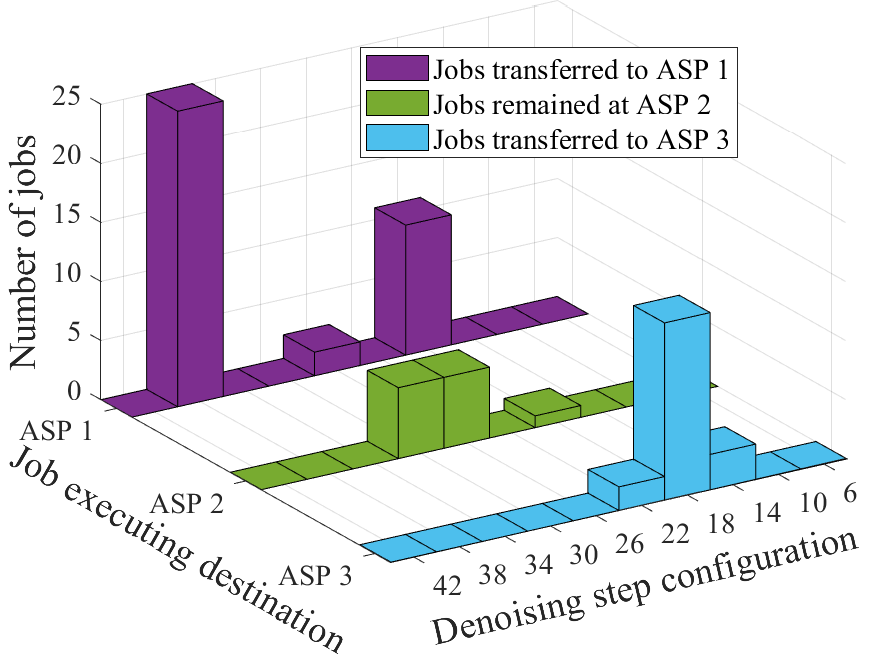}} 
		}
		\vspace{0mm}
		\caption{A snapshot of ASP 2’s job scheduling decisions during 23:00-24:00.}   
		\label{fig:6}     
	\end{figure}
    
    Fig. \ref{fig:6} provides a detailed elaboration on job scheduling decisions of ASP 2 during 23:00-24:00. From Fig. \ref{fig:6} (a), ASP 2 transfers jobs with smaller tolerated delays to ASP 1, while jobs with more relaxed delay constraints are processed locally or offloaded to ASP 3. The rational is that SD1.5 hosted by ASP 1 features a smaller parameter size and shorter execution time, making it more suitable for delay-sensitive tasks, whereas SDXL and SD3.5 can deliver higher AIGC service revenue for delay-tolerant jobs. Fig. \ref{fig:6} (b) shows that the denoising step configurations do not exceed 38, 30, and 22 for the three ASPs, respectively. This observation is consistent with Fig. \ref{fig:2}, where additional steps beyond these thresholds yield marginal revenue gains. It is worth emphasizing that, despite lacking access to other ASPs’ model characteristics, the insights learned through interactive training enable effective distributed and coordinated decision-makings.
    
    \subsection{Learning Performance Evaluation}
    
    \begin{figure}[t] \centering
    	\subfigure[Environmental reward. ] { 
    		{\includegraphics[width=4.1cm]{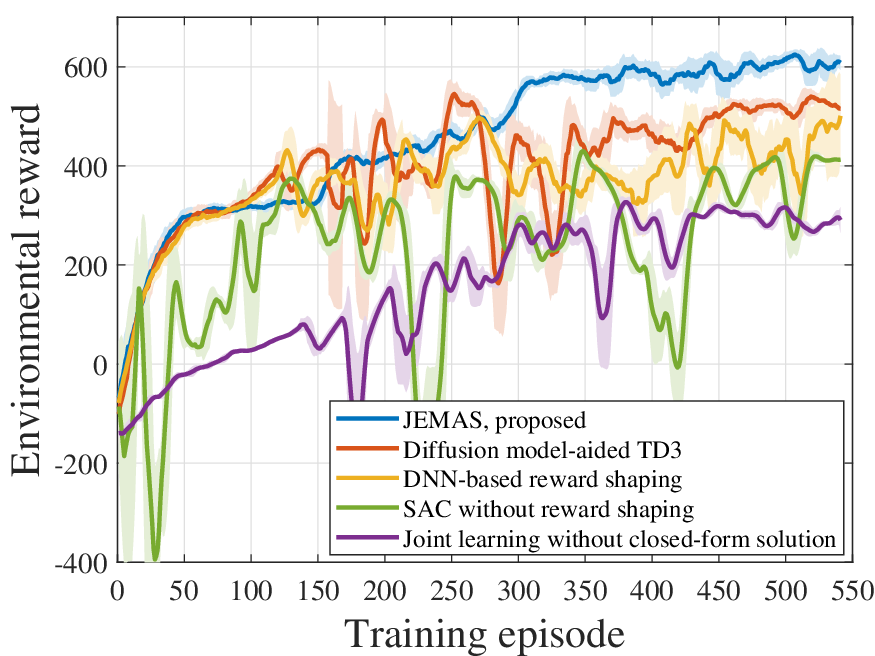}} 
    	}     
    	\subfigure[Delay constraint violation rate.] { 
    		{\includegraphics[width=4.1cm]{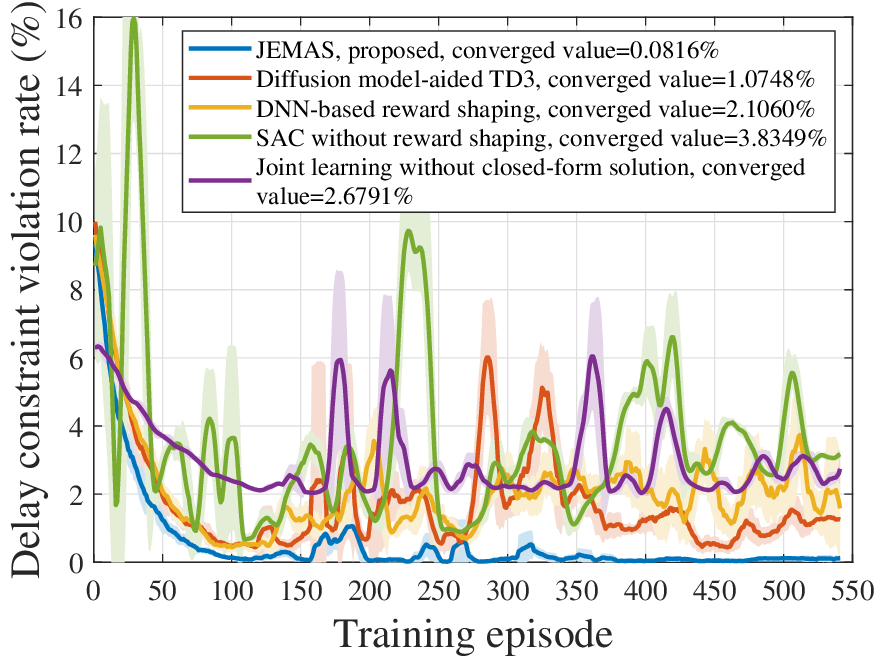}} 
    	}
    	\vspace{0mm}
    	\caption{Learning curves of various DRL schemes.}   
    	\label{fig:7}     
    \end{figure}
    
    \begin{figure}[t] \centering
    	\subfigure[Environmental reward. ] { 
    		{\includegraphics[width=4.1cm]{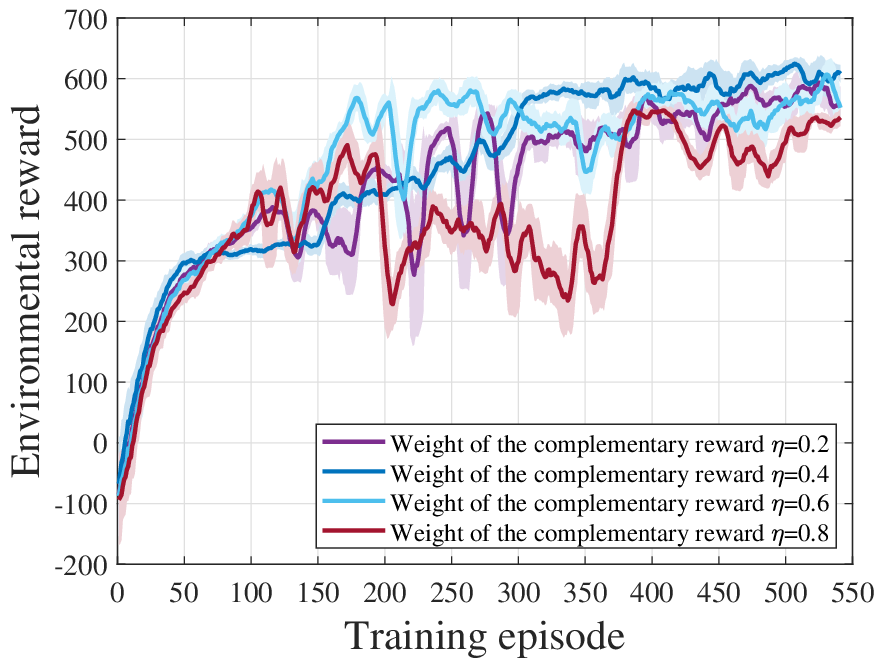}} 
    	}     
    	\subfigure[Generated complementary reward.] { 
    		{\includegraphics[width=4.1cm]{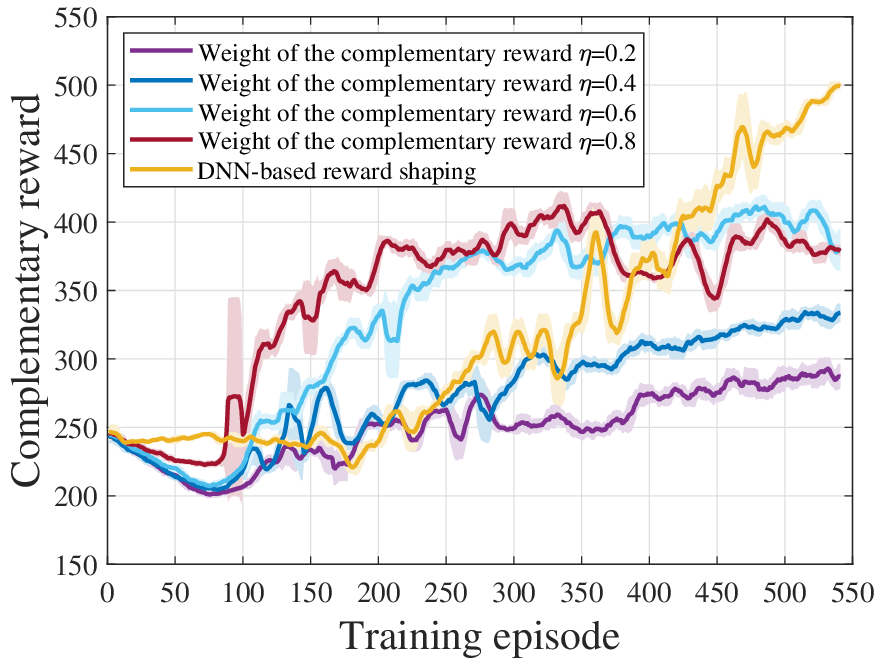}} 
    	}
    	\vspace{0mm}
    	\caption{Learning curves of various reward shaping settings.}   
    	\label{fig:8}     
    \end{figure}
    
    \begin{figure}[t]
    	\begin{tabular}{cc}
    		\begin{minipage}[t]{0.49\linewidth}\centering
    			{\includegraphics[width=4.1cm]{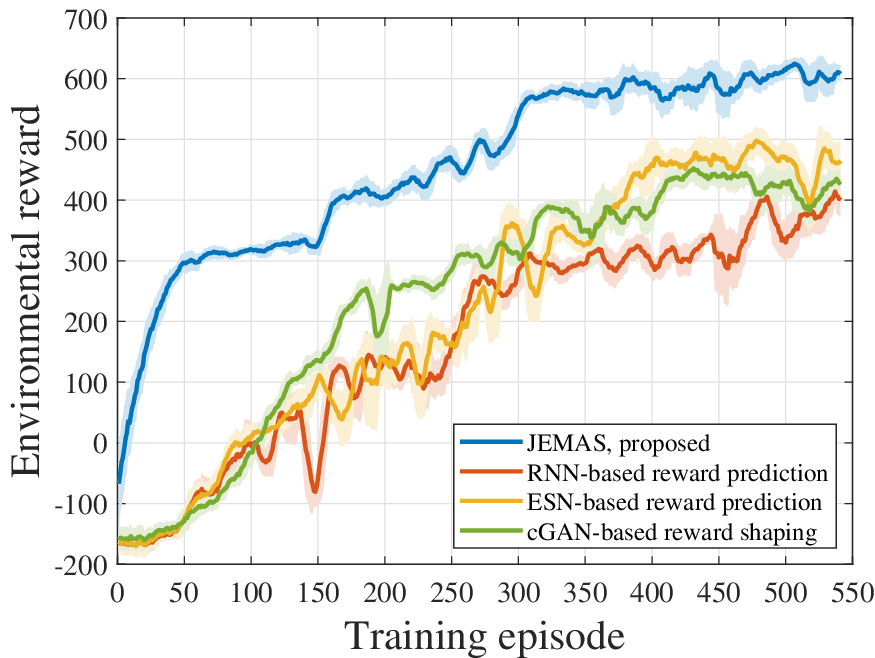}}     
    			\vspace{1mm}
    			\caption{Comparison with reward prediction networks and cGAN model.}    
    			\label{fig:reward_prediction}     
    		\end{minipage}
    		\begin{minipage}[t]{0.49\linewidth}\centering
    			{\includegraphics[width=4.1cm]{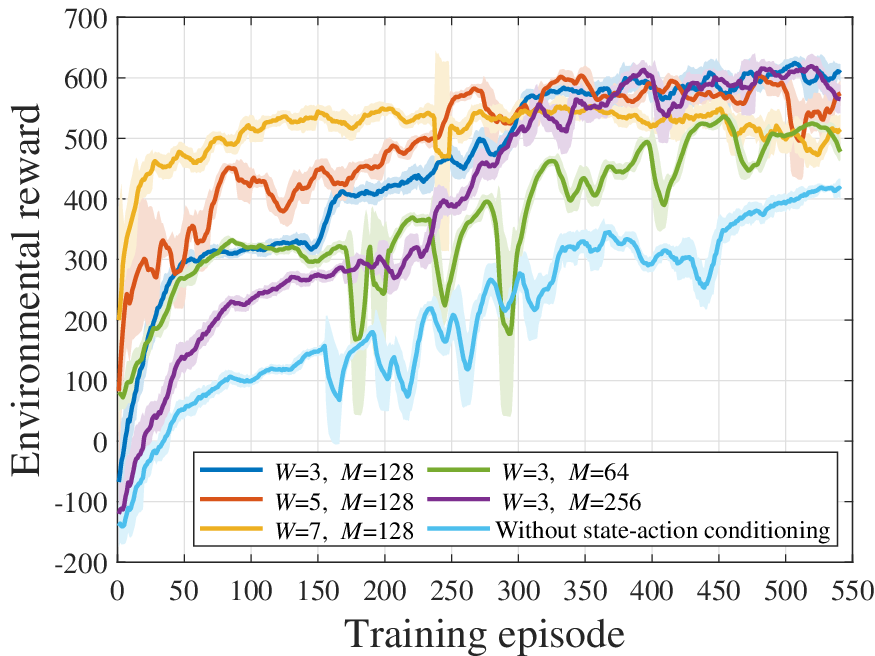}}     
    			\vspace{1mm}
    			\caption{Ablation study for various diffusion model configurations.}    
    			\label{fig:ablation}  
    		\end{minipage}
    		\vspace{0mm}	
    	\end{tabular}
    \end{figure}

    This part evaluates the learning performance of the proposed JEMAS algorithm through a comprehensive ablation study. The benchmark DRL schemes are described as follows. 
    
    \textit{1) Diffusion Model-Aided TD3:} The SAC module in JEMAS is replaced with the twin delayed deep deterministic policy gradient (TD3) algorithm, which also serves to assess the generality of the proposed reward-shaping approach across different DRL paradigms. 
    
    \textit{2) DNN-Based Reward Shaping \cite{36}:} To evaluate the contribution of the diffusion model, this baseline replaces it with a fully connected DNN for generating complementary rewards.
    
    \textit{3) SAC Without Reward Shaping \cite{25}:} To validate the effectiveness of reward shaping, we compare against the standard SAC algorithm without any complementary reward generation. 
     
    \textit{4) Joint Learning Without Closed-Form Solution:} All scheduling and energy management variables are included into the DRL action space and optimized through policy learning.
    
    \textit{5) Various $\eta$ Settings:} The parameter $\eta $ is varied to control the weight of the complementary reward relative to the environmental reward. 

    Fig. \ref{fig:7} (a) plots the average environmental reward (solid curves) and the corresponding standard deviation (shaded regions) received by the DRL agent. Compared with the benchmark schemes, the proposed JEMAS algorithm converges to the highest reward while exhibiting strong learning stability. JEMAS outperforms \textit{diffusion model-aided TD3} by leveraging the stochastic policy updates of SAC, which enable more efficient action exploration than the artificial noise employed in TD3. Moreover, benefiting from the superior distribution modeling and high-quality reward generation capability of diffusion models, JEMAS achieves clear performance gains over \textit{DNN-based reward shaping}. Standard SAC struggles to converge under sparse environmental feedback, underscoring the necessity of reward shaping. Similarly, Fig. \ref{fig:7} (b) showcases that the delay constraint violation rate of JEMAS decreases rapidly during training, with the converged value reduced to 7.59\%, 3.87\%, and 2.13\% of those achieved by the three baselines, respectively. \textit{Joint learning without closed-form solution} suffers from inferior convergence behavior and delay violation rate due to the enlarged action space. This demonstrates the effectiveness of exploiting the problem structure to optimize energy management variables via closed-form solutions while providing reward feedback to DRL. 

    Fig. \ref{fig:8} (a) compares the environmental reward under \textit{various $\eta $ settings}. When $\eta $ is small, the complementary reward is insufficient to effectively guide DRL policy training, whereas an excessively large $\eta $ obscures the true environmental feedback and degrades learning performance. Accordingly, we adopt a proper setting of $\eta =0.4$ in our experiments. As shown in Fig. \ref{fig:7} (b), the complementary reward generated by the diffusion model exhibits a desirable pattern: it initially estimates reward distributions from state-action pairs to facilitate exploration, and subsequently stabilizes to maintain a balance with the environmental reward, thereby promoting exploitation and refining the final performance. In contrast, the complementary reward produced by a fully connected DNN fails to stabilize, leading to action overestimation and hindering the exploitation of policies with high environmental reward. 
    
    Fig. \ref{fig:reward_prediction} compares the learning performance of JEMAS with reward prediction networks, where recurrent neural networks (RNNs) or echo state networks (ESNs) are employed to estimate rewards and generate additional training samples. The results show that JEMAS significantly improves training efficiency compared to conventional reward prediction methods. Furthermore, we consider a conditional generative adversarial network (cGAN) as an alternative generative model for reward shaping. However, it struggles to capture the complex state-action distribution.
    
    Fig. \ref{fig:ablation} presents an ablation study on various diffusion model configurations, including different numbers of denoising steps $W$, latent dimensions $M$, and the use of state-action conditioning. The result shows that increasing $W$ accelerates convergence, whereas an excessively large $W$ weakens the exploration capability of the diffusion model. Increasing $M$ improves the representation capacity for complex environments but incurs higher training complexity. Without state-action conditioning, the diffusion model lacks proper guidance during reward generation, thereby hindering effective training.
    
    \begin{table}[t]\footnotesize
    	\centering
    	\caption{Algorithm Runtime Overhead Evaluation}\label{tab:II}
    	\vspace{2.5 mm}
    	\begin{tabular}{ccc}
    		\Xhline{1\arrayrulewidth}
    		Methods & Training time & Implementation time  \\ 
    		\hline
    		SAC without reward shaping & 45.14 min & 3.801 s \\
    		JEMAS, $W=3$ & 107.31 min & 3.901 s \\
    		JEMAS, $W=5$ & 149.52 min & 3.963 s \\
    		JEMAS, $W=7$ & 215.81 min & 3.993 s \\
    		\Xhline{1\arrayrulewidth}   	
    	\end{tabular}
    \end{table}
    
    Table \ref{tab:II} evaluates the runtime overhead of the proposed JEMAS. Due to the incorporation of diffusion-aided reward shaping for policy learning, JEMAS incurs higher training time than standard SAC, and the runtime overhead increases with the number of denoising steps $W$. Since the diffusion model is not involved after training, all methods exhibit similar implementation time, thereby validating the efficiency of JEMAS in real-time execution.
    
    \subsection{System Utility Evaluation}
    
    \begin{figure*}[t]
    	\begin{tabular}{cccc}
    		\begin{minipage}[t]{0.24\linewidth}\centering
    			{\includegraphics[width=4.1cm]{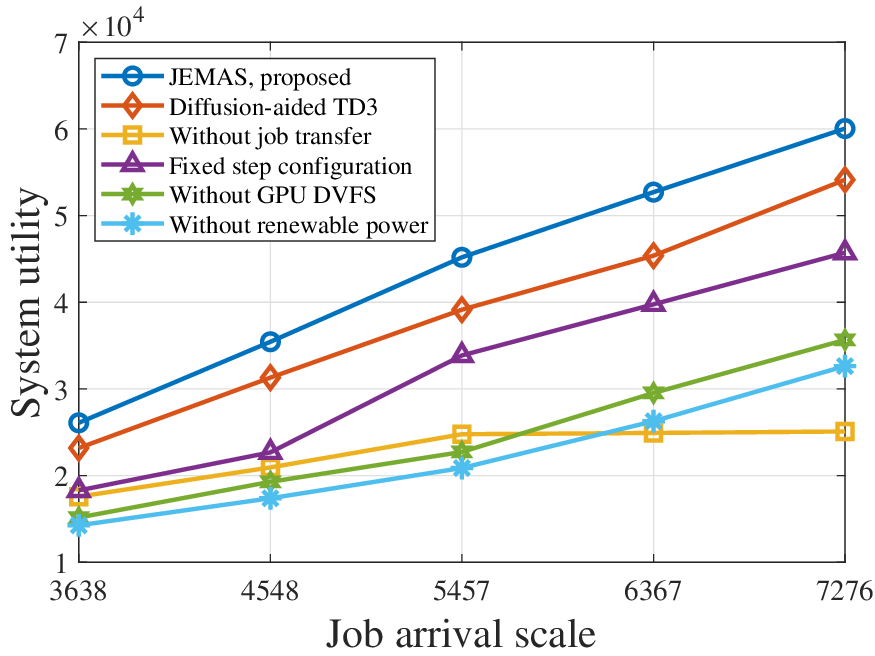}}     
    			\vspace{1mm}
    			\caption{System utility versus job arrival scale.}     
    			\label{fig:9}     
    		\end{minipage}
    		\begin{minipage}[t]{0.24\linewidth}\centering
    			{\includegraphics[width=4.1cm]{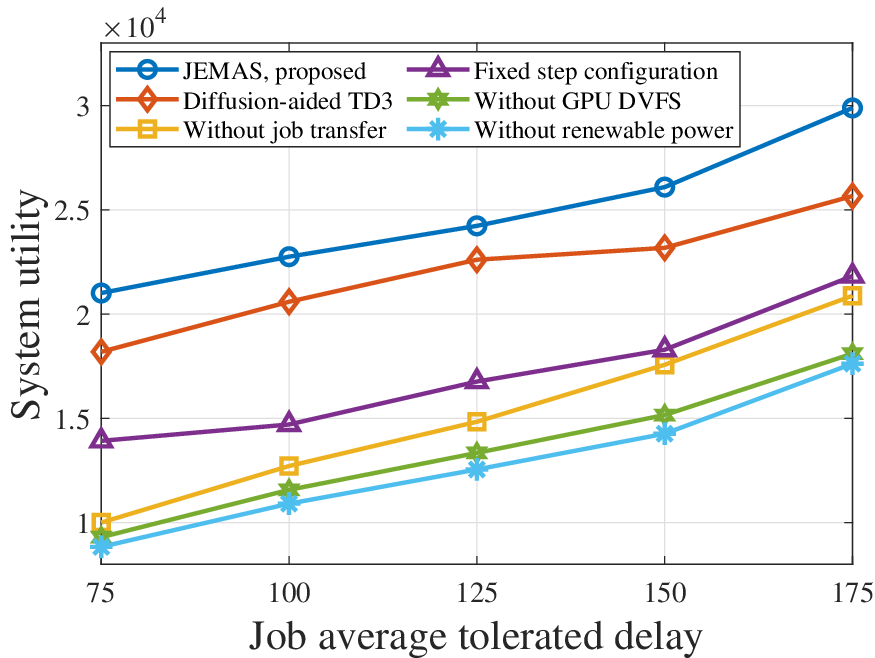}}     
    			\vspace{1mm}
    			\caption{System utility versus job average tolerated delay.}     
    			\label{fig:10}  
    		\end{minipage}
    		\begin{minipage}[t]{0.24\linewidth}\centering
    			{\includegraphics[width=4.1cm]{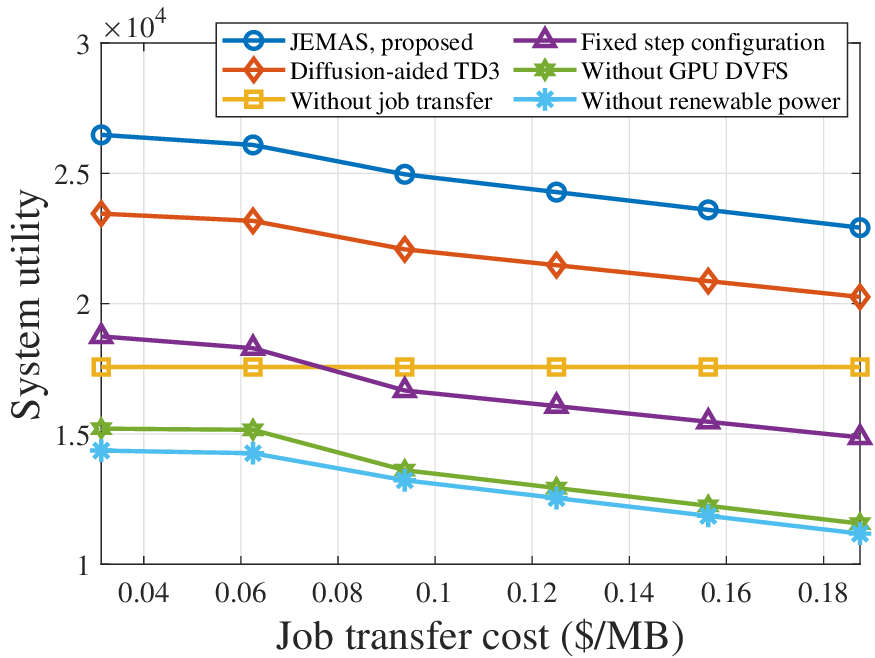}}     
    			\vspace{1mm}
    			\caption{System utility versus job transfer cost.}    
    			\label{fig:11}   
    		\end{minipage}
    		\begin{minipage}[t]{0.24\linewidth}\centering
    			{\includegraphics[width=4.1cm]{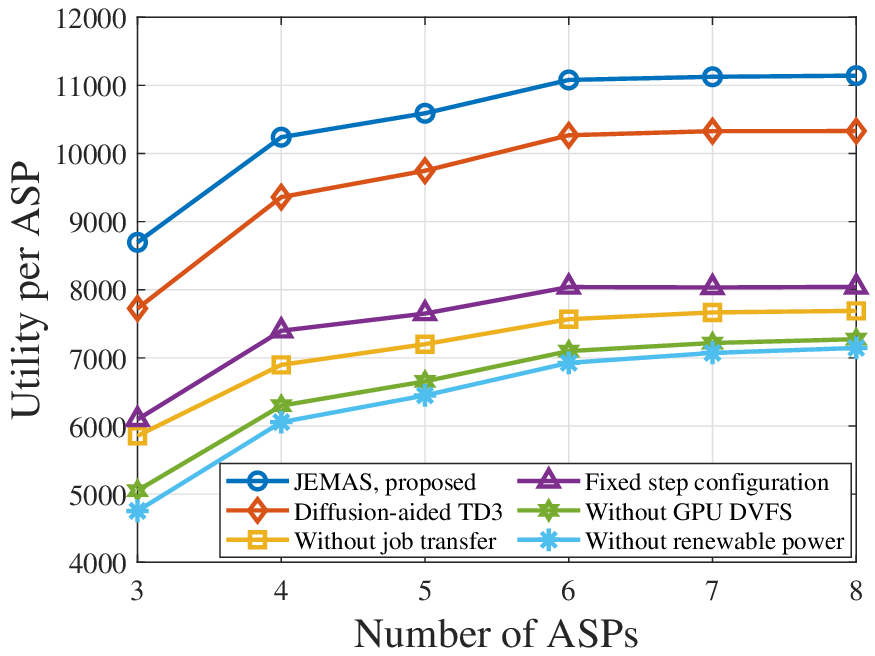}}     
    			\vspace{1mm}
    			\caption{System utility versus number of ASPs.}    
    			\label{fig:scalability}   
    		\end{minipage}
    		\vspace{0mm}	
    	\end{tabular}
    \end{figure*}

    In this subsection, we further assess the system utility realized by JEMAS. In addition to the abovementioned \textit{diffusion-aided TD3}, the following baseline methods are compared. 
    
    \textit{1) Without Job Transfer \cite{18}:} Each job is processed only at its originating ASP, i.e., ${{x}_{n,k}}\left( t \right)=n$, and cooperation among ASPs is prohibited. The denoising step configuration, DVFS and power usage behavior are optimized by our approach. 
    
    \textit{2) Fixed Step Configuration \cite{28}:} The number of denoising steps for each job is fixed as ${{l}_{n,k}}\left( t \right)=30$, and the scheduling decision involves only ASP selection.  
    
    \textit{3) Without GPU DVFS \cite{17}:} The core frequency, core voltage, and memory frequency of GPU servers at all ASPs are fixed at their factory default values, i.e., $\left( f_{n}^{\text{c}}\left( t \right),V_{n}^{\text{c}}\left( t \right),f_{n}^{\text{m}}\left( t \right) \right)=\left( 1,1,1 \right)$. 
    
    \textit{4) Without Renewable Power \cite{15}:} In this worst-case scenario, renewable energy sources are disabled at all ASPs with ${{R}_{n}}\left( t \right)=0$, while other variables are still optimized.

    Fig. \ref{fig:9} shows the relationship between system utility and job arrival scale. As the job scale increases, ASPs can obtain higher AIGC service revenue by successfully processing more jobs, hence most curves exhibit an upward trend. For \textit{without job transfer}, however, many jobs cannot be offloaded to ASPs with lower electricity prices or matched to AIGC models suitable for their delay tolerance and content quality requirements. The resulting increase in energy costs and constraint violation penalties suppresses system utility growth when the job scale exceeds 5457. Moreover, JEMAS outperforms \textit{fixed step configuration} by adaptively selecting denoising steps, thereby enhancing flexibility in balancing service revenue and execution latency. \textit{Without GPU DVFS} performs worse because server power consumption cannot be adjusted to accommodate workload variations, leading to unnecessary energy costs. Comparing JEMAS and \textit{without renewable power} demonstrates that integrating renewable sources at ASPs satisfies a substantial portion of energy demand, significantly improving system utility.

    As shown in Fig. \ref{fig:10}, the system utility increases monotonically with the average tolerated job delay. This is because more jobs can be processed by larger models such as SDXL or SD3.5 and configured with additional denoising steps, thereby enhancing content generation quality and increasing AIGC service revenue. The proposed JEMAS algorithm achieves the highest system utility, yielding improvements of 11.09\%, 38.70\%, 31.03\%, 45.58\%, and 48.24\% over the five baseline methods, respectively. 
    
    Fig. \ref{fig:11} illustrates the impact of job transfer cost on system utility. As the transfer cost increases from 0.0313 to 0.1875 \$/MB, the system utility achieved by JEMAS decreases by 13.45\%. Nevertheless, JEMAS consistently outperforms the benchmark schemes, achieving system utility improvements of 11.47\%, 28.93\%, 32.51\%, 45.59\%, and 47.80\% over the five baselines, respectively.
    
\begin{figure}[t]
    	\begin{tabular}{cc}
    		\begin{minipage}[t]{0.49\linewidth}\centering
    			{\includegraphics[width=4.1cm]{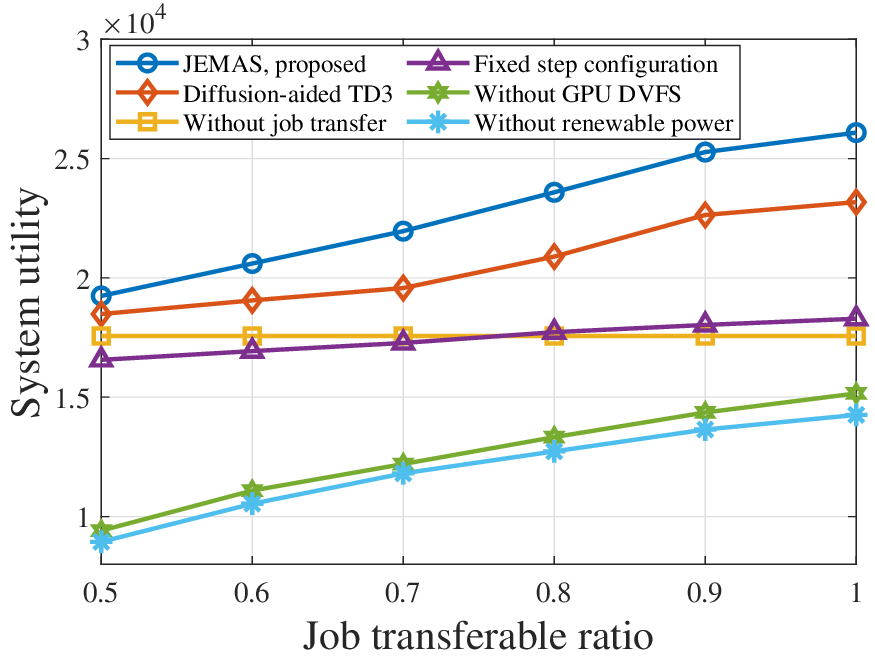}}     
    			\vspace{1mm}
    			\caption{System utility versus job transferable ratio.}   
    			\label{fig:transferable}     
    		\end{minipage}
    		\begin{minipage}[t]{0.49\linewidth}\centering
    			{\includegraphics[width=4.1cm]{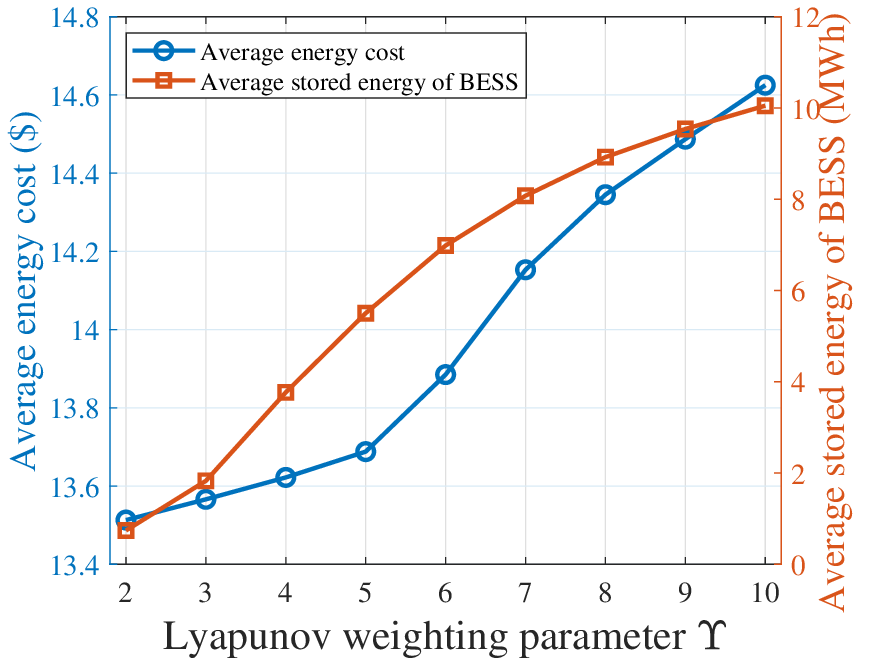}}     
    			\vspace{1mm}
    			\caption{Impact of Lyapunov weighting parameter $\Upsilon$.}    
    			\label{fig:Lyapunov}  
    		\end{minipage}
    		\vspace{0mm}	
    	\end{tabular}
    \end{figure}

    Fig. \ref{fig:scalability} shows that as the number of ASPs $N$ increases, the utility per ASP first rises and then gradually saturates. While additional ASP resources allow more jobs to be completed, the accompanying increase in energy utilization and costs limits further utility improvement. Moreover, Fig. \ref{fig:transferable} illustrates that the utility decreases as more jobs become nontransferable, owing to the reduced degrees of freedom in scheduling. Particularly, JEMAS consistently outperforms all baseline algorithms across different values of $N$ and job transferable ratios, thereby validating the realism and scalability of the proposed approach.
    
    Fig. \ref{fig:Lyapunov} examines the impact of Lyapunov weighting parameter $\Upsilon$. As $\Upsilon$ increases, JEMAS places greater emphasis on BESS energy storage, leading to higher energy cost for coping with future uncertainties over a long time horizon.
    
    \begin{figure}[t]\centering
    	\includegraphics[width = 8cm]{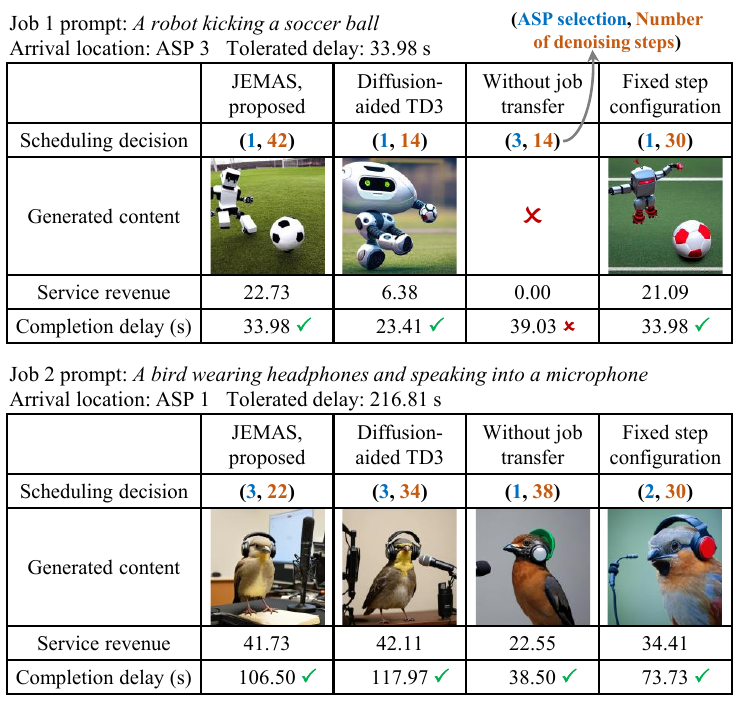}
    	\vspace{1mm}
    	\caption{Scheduling decisions and generation results of various methods for two jobs arriving at 16:05.} 
    	\label{fig:12}   
    \end{figure}
    
    Fig. \ref{fig:12} compares different methods in terms of scheduling decisions and generation results for two jobs arriving at 16:05. For both the delay-sensitive Job 1 and delay-tolerant Job 2, JEMAS assigns them to the most suitable ASP with appropriate denoising step configurations, achieving high-quality image generation under satisfactory latency and low energy cost. These results highlight the applicability of JEMAS as a promising scheduling framework for future large-scale cloud AIGC deployments. 
    
	\section{Conclusion} \label{sec:conclusion}	
	
	In this work, we developed a joint energy management and coordinated AIGC workload scheduling framework for distributed data centers, aiming to deliver high-quality AIGC services while reducing operational costs. Within this framework, job scheduling (including ASP selection and denoising step configuration) are jointly optimized with GPU DVFS and power usage behaviors to maximize system utility that accounts for AIGC service revenue, deadline violation penalties, job transfer costs, and energy costs. To address the challenges posed by distributed decision-making and reward sparsity, we proposed JEMAS, a diffusion model-enhanced DRL approach that enables coordinated AIGC workload scheduling without requiring access to the private model information of ASPs. Specifically, JEMAS conditions the denoising process on state-action pairs to synthesize complementary reward signals, which are leveraged to guide the SAC agent in learning effective scheduling policies under sparse environmental feedback. Besides, JEMAS incorporates an efficient heuristic with closed-form solutions to optimize GPU DVFS and power usage behaviors. Comprehensive experiments based on real-world AIGC models and datasets demonstrated that JEMAS effectively assigns jobs to suitable ASPs with appropriately configured denoising steps, while adapting to electricity price fluctuations and AIGC model heterogeneity. Compared with benchmark methods, JEMAS significantly accelerates learning convergence and improves system utility by more than 30\%. 
	
	\begin{appendices} 
		\section{Proof of Theorem 1}
		
		After a close observation of \textbf{SP2.1}$_{n}\left( t \right)$, we notice that when $\bm{\alpha} \left( t \right)$ is fixed, the target of DVFS is equivalent to minimizing the GPU power consumption $P_{n}^{\text{G}}\left( t \right)$ while guaranteeing the delay constraints. The rational is that a smaller $P_{n}^{\text{G}}\left( t \right)$ alleviates the burden of the cooling system and the BESS, diminishing power absorption from the grid and thus reducing energy cost. Therefore, we recast the problem for optimizing ${{\mathbf{f}}_{n}}\left( t \right)$ as follows
		\begin{subequations}
			\begin{align}
				\underset{{{\mathbf{f}}_{n}}\left( t \right)}{\mathop{\min }}\,P_{n}^{\text{G}}\left( t \right)=P_{n}^{0}+{{\lambda }_{n}}f_{n}^{\text{m}}\left( t \right)+{{\delta }_{n}}{{\left( V_{n}^{\text{c}}\left( t \right) \right)}^{2}}f_{n}^{\text{c}}\left( t \right), \label{eq:41a}
			\end{align}
			\vspace{-5mm}
			\begin{align}
				\mbox{s.t.}\ 
				&g\left( {{\mathbf{f}}_{n}}\left( t \right) \right)\le {{\tilde{d}}^{\min }}\left( t \right), \label{eq:41b}\\
				&V_{n}^{\text{c},\min }\le V_{n}^{\text{c}}\left( t \right)\le V_{n}^{\text{c},\max }, \label{eq:41c}\\
				&f_{n}^{\text{c},\min }\le f_{n}^{\text{c}}\left( t \right)\le h\left[ V_{n}^{\text{c}}\left( t \right) \right], \label{eq:41d}\\
				&f_{n}^{\text{m},\min }\le f_{n}^{\text{m}}\left( t \right)\le f_{n}^{\text{m},\max }, \label{eq:41e}
			\end{align} 
		\end{subequations}
		where (\ref{eq:41b}) is rewritten from the delay constraint (\ref{eq:36b}), with $g\left( {{\mathbf{f}}_{n}}\left( t \right) \right)=\Delta _{n}^{0}+{{\varepsilon }_{n}}\left( \frac{{{\kappa }_{n}}}{f_{n}^{\text{c}}\left( t \right)}+\frac{1-{{\kappa }_{n}}}{f_{n}^{\text{m}}\left( t \right)} \right)$, ${{\tilde{d}}_{k}}\left( t \right)=\frac{2{{\Theta }_{n}}\left[ {{d}_{k}}\left( t \right)-\Delta _{k}^{\text{tra}}\left( t \right)-\Delta _{k}^{\text{fee}}\left( t \right) \right]}{\sum\nolimits_{{k}'\in \mathcal{K}_{n}^{*}\left( t \right)}{{{l}_{{{k}'}}}\left( t \right)}+{{l}_{k}}\left( t \right)}$, and ${{\tilde{d}}^{\min }}\left( t \right)$ is the minimum delay requirement among all ${{\tilde{d}}_{k}}\left( t \right)$ with ${{\alpha }_{k}}\left( t \right)=1$. As $P_{n}^{\text{G}}\left( t \right)$ monotonically increases with ${{\mathbf{f}}_{n}}\left( t \right)$, $g\left( {{\mathbf{f}}_{n}}\left( t \right) \right)={{\tilde{d}}^{\min }}\left( t \right)$ must hold at the optimal solution, thus we have 
		\begin{align}
			 f_{n}^{\text{m}}\left( t \right)=\frac{{{\varepsilon }_{n}}\left( 1-{{\kappa }_{n}} \right)}{{{{\tilde{d}}}^{\min }}\left( t \right)-\Delta _{n}^{0}-\frac{{{\varepsilon }_{n}}{{\kappa }_{n}}}{f_{n}^{\text{c}}\left( t \right)}}. \label{eq:42}
		\end{align} 
	    By substituting (\ref{eq:42}) into (\ref{eq:41a}), we calculate the first-order derivative of $P_{n}^{\text{G}}\left( t \right)$ with respect to (w.r.t.) $f_{n}^{\text{c}}\left( t \right)$ as
	    \begin{align}
	    	\frac{\partial P_{n}^{\text{G}}\!\left( t \right)}{\partial f_{n}^{\text{c}}\left( t \right)}\!=\!\frac{-{{\lambda }_{n}}{{\varepsilon }_{n}}\left( 1-{{\kappa }_{n}} \right){{\varepsilon }_{n}}{{\kappa }_{n}}}{{{\{\! [ {{{\tilde{d}}}^{\min }}\!\left( t \right)\!-\!\Delta _{n}^{0} ] f_{n}^{\text{c}}\left( t \right)\!-\!{{\varepsilon }_{n}}{{\kappa }_{n}} \!\}\!}^{2}}}\!\!+\!{{\delta }_{n}}{{\left( V_{n}^{\text{c}}\!\left( t \right) \right)}^{2}},\!\! \label{eq:43}
	    \end{align} 
	    Let $\frac{\partial P_{n}^{\text{G}}\left( t \right)}{\partial f_{n}^{\text{c}}\left( t \right)}=0$, we derive the stationary point as 
	    \begin{align}
	    	\tilde{f}_{n}^{\text{c}}\left( t \right)=\frac{{{\varepsilon }_{n}}{{\kappa }_{n}}}{{{{\tilde{d}}}^{\min }}\left( t \right)-\Delta _{n}^{0}}+\frac{{{\varepsilon }_{n}}\sqrt{{{{\lambda }_{n}}\left( 1-{{\kappa }_{n}} \right){{\kappa }_{n}}}/{{{\delta }_{n}}}}}{[ {{{\tilde{d}}}^{\min }}\left( t \right)-\Delta _{n}^{0} ]V_{n}^{\text{c}}\left( t \right)}. \label{eq:44}
	    \end{align} 
        Since the second-order derivate $\frac{{{\partial }^{2}}P_{n}^{\text{G}}\left( t \right)}{\partial {{\left[ f_{n}^{\text{c}}\left( t \right) \right]}^{2}}}=\frac{2{{\lambda }_{n}}{{\varepsilon }_{n}}\left( 1-{{\kappa }_{n}} \right){{\varepsilon }_{n}}{{\kappa }_{n}}\left[ {{{\tilde{d}}}^{\min }}\left( t \right)-\Delta _{n}^{0} \right]}{{{\left\{ \left[ {{{\tilde{d}}}^{\min }}\left( t \right)-\Delta _{n}^{0} \right]f_{n}^{\text{c}}\left( t \right)-{{\varepsilon }_{n}}{{\kappa }_{n}} \right\}}^{2}}}>0$, $P_{n}^{\text{G}}\left( t \right)$ is a convex function w.r.t. $f_{n}^{\text{c}}\left( t \right)$, thus the optimal $f_{n}^{\text{c}}\left( t \right)$ is expressed as 
        \begin{align}
        	f_{n}^{\text{c*}}\left( t \right)=
        	\left\{\begin{array}{cl}
        		&\!\!\!\!\!\!\!\!\! f_{n}^{\text{c},\min },\ \tilde{f}_{n}^{\text{c}}\left( t \right)\le f_{n}^{\text{c},\min },\\
        		&\!\!\!\!\!\!\!\!\!\tilde{f}_{n}^{\text{c}}\left( t \right),\ f_{n}^{\text{c},\min }<\tilde{f}_{n}^{\text{c}}\left( t \right)\le h\left[ V_{n}^{\text{c}}\left( t \right) \right],\\
        		&\!\!\!\!\!\!\!\!\! h\left[ V_{n}^{\text{c}}\left( t \right) \right],\ \tilde{f}_{n}^{\text{c}}\left( t \right)>h\left[ V_{n}^{\text{c}}\left( t \right) \right].
        	\end{array}\right.  \label{eq:45}
        \end{align}
        Accordingly, we conduct a simply one-dimensional search of $V_{n}^{\text{c}}\left( t \right)$ over $\left[ V_{n}^{\text{c},\min },V_{n}^{\text{c},\max } \right]$, then calculate $f_{n}^{\text{c*}}\left( t \right)$ and $f_{n}^{\text{m*}}\left( t \right)$ (which should be greater than $f_{n}^{\text{m},\min }$) based on (\ref{eq:45}) and (\ref{eq:42}), respectively, thereby finding the optimal $V_{n}^{\text{c*}}\left( t \right)$ that minimizes $P_{n}^{\text{G}}\left( t \right)$. If there is no $V_{n}^{\text{c}}\left( t \right)$ such that $f_{n}^{\text{m*}}\left( t \right)$ satisfies (\ref{eq:41e}), the problem is infeasible. 
        
        After obtaining $P_{n}^{\text{G*}}\left( t \right)$ by substituting $\mathbf{f}_{n}^{*}\left( t \right)$ into (\ref{eq:41a}), we also minimize $P_{n}^{\text{C}}\left( t \right)$ of the cooling system to reduce energy cost while ensuring the data center temperature constraint (\ref{eq:23}). This is given by
        \begin{align}
        	P_{n}^{\text{C*}}\left( t \right)=-\min \left\{ \frac{\tilde{\zeta }_{n}^{\max }\left( t \right)}{{{\vartheta }^{\text{COP}}}},\frac{\Delta \tilde{\zeta }_{n}^{\max }\left( t \right)}{{{\vartheta }^{\text{COP}}}} \right\}+\frac{P_{n}^{\text{G*}}\left( t \right)}{{{\vartheta }^{\text{COP}}}}, \label{eq:46}
        \end{align} 
        and the problem is infeasible when $P_{n}^{\text{C*}}\left( t \right)>P_{n}^{\text{C},\max }$. 
        
        For BESS discharging power ${{D}_{n}}\left( t \right)$, as all other variables are determined, we can rewrite \textbf{SP2.1}$_{n}\left( t \right)$ in (\ref{eq:36}) as 
        \begin{subequations}
        	\begin{align}
        		\underset{{{D}_{n}}\left( t \right)}{\mathop{\max }}\,\left[ {{\varsigma }_{n}}\left( t \right)-\Upsilon {{{\tilde{E}}}_{n}}\left( t \right) \right]{{D}_{n}}\left( t \right), \label{eq:47a}
        	\end{align}
        	\vspace{-5mm}
        	\begin{align}
        		\mbox{s.t.}\ 
        		-\!\min \{ D_{n}^{\text{min}},{{{{\tilde{E}}}_{n}}\!\left( t \right)}/{\tau } \}\!\le\! {{D}_{n}}\!\left( t \right)\!\le\! \min \{\! D_{n}^{\text{max}},{{{E}_{n}}\!\left( t \right)}/{\tau } \}. \label{eq:47b}
        	\end{align} 
        \end{subequations}
        Obviously, (\ref{eq:47a}) is linear w.r.t. ${{D}_{n}}\left( t \right)$, the optimal solution follows the threshold-based structure in the sequel
        \begin{align}
        	D_{n}^{*}\left( t \right)\!=\!
        	\left\{\begin{array}{cl}
        		&\!\!\!\!\!\!\!\!\! -\!\min \{ D_{n}^{\text{min}},{{{{\tilde{E}}}_{n}}\left( t \right)}/{\tau }\},{{\varsigma }_{n}}\left( t \right)\le \Upsilon {{{\tilde{E}}}_{n}}\left( t \right),\\
        		&\!\!\!\!\!\!\!\!\! \min \{ D_{n}^{\text{max}},{{{E}_{n}}\left( t \right)}/{\tau } \},\ \ {{\varsigma }_{n}}\left( t \right)>\Upsilon {{{\tilde{E}}}_{n}}\left( t \right).
        	\end{array}\right.  \label{eq:48}
        \end{align}
        This completes the proof. 
        
        \section{Proof of Theorem 2}
        To theoretically verify that diffusion-aided reward shaping remains equivalent to maximizing the system utility, we first prove that a potential-based shaping function is necessary for guaranteeing consistency with the optimal policy \cite{37}. On this basis, we show that the complementary reward generated by the diffusion model is optimized to align with the potential-based shaping structure. 
        
        Specifically, denote the original MDP for system utility maximization and the MDP after reward shaping as $\mathcal{M}$ and $\mathcal{M}'$, respectively. For $\mathcal{M}$, its optimal Q-function $Q^*_\mathcal{M}$ satisfies the Bellman equation, i.e.,
        \begin{align}
        	Q^*_\mathcal{M}(\mathbf{o},\mathbf{a})=\mathbb{E}[r^{\text{E}}+\gamma \underset{\mathbf{\bar{a}}}{\max} Q^*_\mathcal{M}(\mathbf{\bar{o}},\mathbf{\bar{a}})]. \label{eq:A1}
        \end{align}
        We can transform it into
        \begin{align}
        	&Q^*_\mathcal{M}(\mathbf{o},\mathbf{a})-\Lambda(\mathbf{o},\mathbf{a})=\mathbb{E}[r^{\text{E}}+\Gamma\Lambda(\mathbf{\bar{o}},\mathbf{\bar{a}})-\Lambda(\mathbf{o},\mathbf{a}) \nonumber\\
        	&\qquad\qquad\qquad\ +\Gamma\underset{\mathbf{\bar{a}}}{\max} (Q^*_\mathcal{M}(\mathbf{\bar{o}},\mathbf{\bar{a}})-\Lambda(\mathbf{\bar{o}},\mathbf{\bar{a}}))], \label{eq:A2}
        \end{align}
        where $\Lambda(\mathbf{o},\mathbf{a})$ represent the potential function over state $\mathbf{o}$ and action $\mathbf{a}$. Define $\eta\cdot r^\text{C}\triangleq \Gamma\Lambda(\mathbf{\bar{o}},\mathbf{\bar{a}})-\Lambda(\mathbf{o},\mathbf{a})$ as the complementary reward, and $Q_{\mathcal{M}'}(\mathbf{o},\mathbf{a})\triangleq Q^*_\mathcal{M}(\mathbf{o},\mathbf{a})-\Lambda(\mathbf{o},\mathbf{a})$ as the Q-function for $\mathcal{M}'$, then we have
        \begin{align}
        	Q_{\mathcal{M}'}(\mathbf{o},\mathbf{a})=\mathbb{E}[r^{\text{E}}+\eta\cdot r^{\text{C}}+\gamma \underset{\mathbf{\bar{a}}}{\max}Q_{\mathcal{M}'}(\mathbf{\bar{o}},\mathbf{\bar{a}})]. \label{eq:A3}
        \end{align}
        Since $r^{\text{E}}+\eta\cdot r^{\text{C}}$ is the total reward received by the agent according to our reward shaping approach, the above formula is exactly the Bellman equation for $\mathcal{M}'$. Therefore, the optimal Q-function is given by $Q^*_{\mathcal{M}'}(\mathbf{o},\mathbf{a})=Q^*_\mathcal{M}(\mathbf{o},\mathbf{a})-\Lambda(\mathbf{o},\mathbf{a})$. In addition, the optimal policy for $\mathcal{M}'$ can be expressed as
        \begin{align}
        	\pi_{\mathcal{M}'}(\mathbf{a}|\mathbf{o})&\in \arg\underset{\mathbf{a}}{\max}Q^*_{\mathcal{M}'}(\mathbf{o},\mathbf{a})\nonumber\\
        	&=\arg\underset{\mathbf{a}}{\max}Q^*_\mathcal{M}(\mathbf{o},\mathbf{a})-\Lambda(\mathbf{o},\mathbf{a})\nonumber\\
        	&=\arg\underset{\mathbf{a}}{\max}Q^*_\mathcal{M}(\mathbf{o},\mathbf{a})=\pi_{\mathcal{M}}(\mathbf{a}|\mathbf{o}). \label{eq:A4}
        \end{align}
        This demonstrates that the optimal policies for MDPs $\mathcal{M}$ and $\mathcal{M}'$ are equivalent, hence adding complementary rewards following the potential-based shaping structure $\Gamma\Lambda(\mathbf{\bar{o}},\mathbf{\bar{a}})-\Lambda(\mathbf{o},\mathbf{a})$ does not impact the optimality. 
        
        However, designing an explicit potential function $\Lambda$ is non-trivial in AIGC job scheduling environments with complex state-action spaces. In this work, we employ a diffusion model to learn the complementary rewards, which implicitly captures the latent structure of environmental rewards. Subsequently, we elucidate that the diffusion model is optimized to align with the potential-based shaping structure. In the proposed approach, $\bm{\theta}$ is trained under the supervision of the evaluation network ${Y}_{\bm{\varphi}}$, and ${Y}_{\bm{\varphi}}$ is updated to minimize $L\left( \bm{\varphi}  \right)$. When $L\left( \bm{\varphi}  \right)=0$, we have
        \begin{align}
        	r_{n,k}^{\text{E}}\left( t \right)\!=\!-\Gamma {{Y}_{{\hat{\bm{\varphi} }}}}\left( {{{\mathbf{\bar{v}}}}_{n,k}}\left( t \right),\bar{r}_{n,k}^{\text{C}}\left( t \right) \right)\!+\!{{Y}_{\bm{\varphi} }}\left( {{\mathbf{v}}_{n,k}}\left( t \right),r_{n,k}^{\text{C}}\left( t \right) \right), \label{eq:A5}
        \end{align}
        while $r_{n,k}^{\text{C}}\left( t \right)$ is learned to approximate $r_{n,k}^{\text{E}}\left( t \right)$ through gradient descent. By treating $-{{Y}_{\bm{\varphi} }}( {{\mathbf{v}}_{n,k}}\left( t \right),r_{n,k}^{\text{C}}\left( t \right) )$ as a kind of potential function, $r_{n,k}^{\text{C}}\left( t \right)$ generated by the diffusion model follows a consistent structure with the optimal potential-based reward shaping. Recall that such structure maintains the optimal policy of the original MDP, thus the proposed diffusion-aided reward shaping remains equivalent to maximizing the system utility. This completes the proof.
        
	\end{appendices}
     
	\bibliographystyle{IEEEtran}
	\bibliography{reference}
   
\end{document}